\documentclass{article} 

\usepackage{arxiv}
\usepackage{graphicx}
\usepackage{amsmath,amssymb,amsfonts, epsf}
\usepackage{algorithmic}
\usepackage[linesnumbered, ruled, vlined]{algorithm2e}
\usepackage{lineno,hyperref}
\usepackage{breqn}
\graphicspath{{images/}}
\usepackage{nomencl}
\hypersetup{
    colorlinks=true,
    linkcolor=blue,
    filecolor=magenta,      
    urlcolor=cyan,
}
\usepackage{xcolor}
\usepackage{caption}

\title{Implicit supervision for fault detection and segmentation of emerging fault types with Deep Variational Autoencoders}

\author{
  Manuel Arias Chao \\
  ETH Zurich\\
  \texttt{manuel.arias@ethz.ch} \\
   \And
  Bryan T. Adey \\
  ETH Zurich\\
  \texttt{adey@ibi.baug.ethz.ch} \\
   \And
  Olga Fink \\
  ETH Zurich\\
  \texttt{ofink@ethz.ch} \\
}

\begin{document}
\maketitle

\begin{abstract}
Data-driven fault diagnostics of safety-critical systems often faces the challenge of a complete lack of labeled data associated with faulty system conditions (i.e., fault types) at training time. Since an unknown number and nature of fault types can arise during deployment, data-driven fault diagnostics in this scenario is an \textit{open-set} learning problem. Most of the algorithms for open-set diagnostics are one-class classification and unsupervised algorithms that do not leverage all the available labeled and unlabeled data in the learning algorithm. As a result, their fault detection and segmentation performance (i.e., identifying and separating faults of different types) are sub-optimal. With this work, we propose training a variational autoencoder (VAE) with labeled and unlabeled samples while inducing implicit supervision on the latent representation of the healthy conditions. This, together with a modified sampling process of VAE, creates a compact and informative latent representation that allows good detection and segmentation of unseen fault types using existing one-class and clustering algorithms. We refer to the proposed methodology as "knowledge induced variational autoencoder with adaptive sampling" (\text{KIL-AdaVAE}). The fault detection and segmentation capabilities of the proposed methodology are demonstrated in a new simulated case study using the Advanced Geared Turbofan 30000 (AGTF30) dynamical model under real flight conditions. In an extensive comparison, we demonstrate that the proposed method outperforms other learning strategies (supervised learning, supervised learning with embedding and semi-supervised learning) and deep learning algorithms, yielding significant performance improvements on fault detection and fault segmentation.
\end{abstract}


\section{Introduction}
% Introduction: 1) social importance --> Faults in complex systems
Proper functioning of our society depends on complex industrial systems to provide an adequate level of service. Although critical faults in complex industrial systems are rare, they can cause substantial economic and social disruptions if they occur. To mitigate the impact of faults, they need to be detected early (i.e., fault detection). To appropriately plan maintenance interventions, the different fault types need to be distinguished (i.e., fault diagnostics).

% Introduction: 2) the sub-field --> Data driven diagnostics
The recent increased availability of condition monitoring (CM) data has triggered the use of data-driven fault diagnostics methods on complex industrial systems. The accuracy of these methods is generally satisfactory when two conditions are met: 1) abundant labeled data on healthy and faulty system conditions exists, and 2) the diagnosis problem is formulated as a supervised classification task (e.g., \cite{Yin2016, Jing2017, Wang2018StackedSA, Lei2019}). In real world applications, however, it is common that only a small fraction of the system CM data are labeled as healthy at training time. The rest is unlabeled due to the uncertainty of the number and type of faults that may occur. Under these circumstances, supervised fault diagnostics models often perform poorly.

\begin{figure}[ht]
\centering
\includegraphics[width=9.7cm]{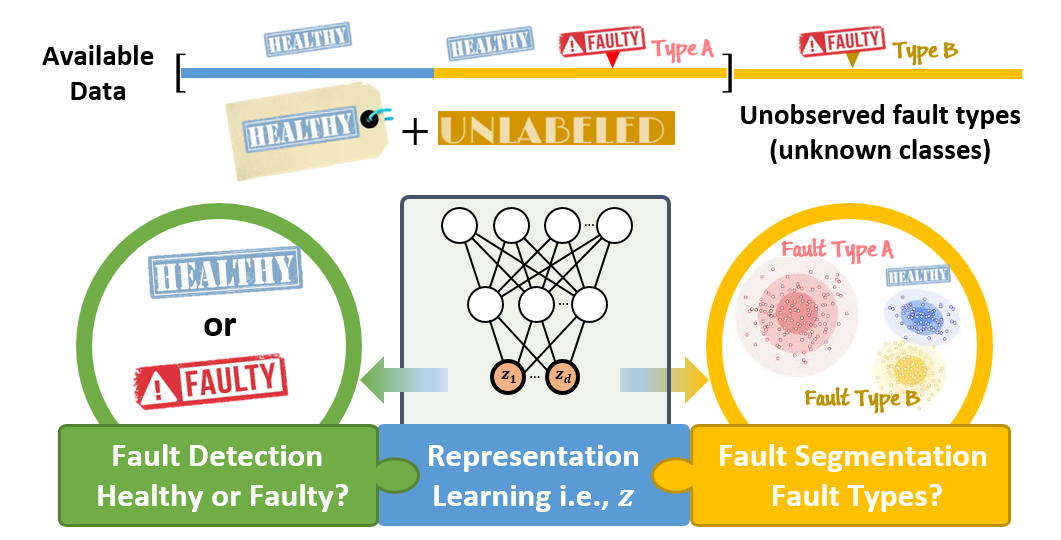} 
    \caption{Open-set diagnostics and proposed solution strategy. The available data at training time does not contain samples of all possible fault types, and in many cases, it may only consist of healthy labeled data. \textbf{Fault detection} aims at detecting faulty system conditions. \textbf{Fault segmentation} aims at partitioning the system conditions into disjoint sets of healthy system conditions and system conditions affected by different fault types. In this work, a representation learning algorithm is proposed to obtain a compact and informative latent representation (i.e., $z$) that allows good detection and segmentation of unseen fault types using existing one-class and density-based clustering algorithms.}
    \label{fig:openset}
\end{figure}

% Introduction: 3) the specify context: open-set diagnostics
Data-driven fault diagnostics with an unknown number and nature of faults at model development time is an \textit{open-set} learning problem \cite{scheirer2012toward}. Since the datasets contain system conditions that have not been observed before, open-set problems are harder than those with a complete knowledge of all fault types. Unlike classical fault diagnostics that is typically solved as a classification task, fault detection and diagnostics are typically addressed independently in open-set problems. Moreover, open-set fault diagnostics comprises an unsupervised fault segmentation task that aims to identify and separate faults of different types (see Figure \ref{fig:openset}).

% The problem: existing method are not tailored to the open-set scenario
In open-set scenarios, fault detection is often formulated as an anomaly detection (AD) task. The aim of the AD task is to learn a model that accurately describes "healthy" conditions. Deviations from the learned patterns are then considered to be anomalies. One of the common approaches for AD is one-class classification \cite{Moya1996, Scholkopf2000, pmlr-v80-ruff18a, chalapathy2018anomaly}. Examples of fault detection solutions based on ones-class classifiers include \cite{Fernandez-Francos2013, Michau2017, AriasChao2019}. These and other alternative AD methods \cite{Song2017AHS, Akay2018GANomalySA, Chalapathy2019} are, however, designed to only use healthy labeled data in the learning algorithm. This means that when they are applied to an open-set scenario, the unlabeled data are not used. The opposite situation is present in fault segmentation of fault types that have not been seen in the training dataset. Clustering methods directly applied to the raw CM data \cite{Detroja2006, HuDi2012, Li2014, Du2014a, Costa2015, Zhu2017} are the most commonly used fault segmentation methods. Clustering methods are, however, generally unsupervised methods. Therefore, labels of healthy system conditions are not part of the learning algorithm. In both cases, existing methods do not leverage the available data in the open-set diagnostics scenario, restricting their learning potential.

% Some light --> a possible way forward but not quite
An intuitive way to incorporate the labeled and unlabeled data in the learning algorithm is to formulate the problem as semi-supervised learning (SSL) problem. Concretely, SSL methods use structural assumptions to leverage unlabeled data in the main supervised task \cite{Ratner2019}. Most of the semi-supervised methods for diagnostics \cite{Jiang2013, Hu2017, Chen2018, Razavi-Far2019} focus on feature extraction for multi-class classification problems with a training dataset representative of all the possible fault types (i.e., each fault type is a class). Such semi-supervised classifiers typically assume that similar points are likely to be of the same class; this is known as the cluster assumption \cite{Chapelle}. As highlighted in \cite{Ruff2020Deep}, this assumption, however, only holds for the "normal class" in an AD formulation. For the "anomalous class," the cluster assumption is invalid since anomalies are not necessarily similar. For this reason, a standard SSL AD formulation can be a useful detection strategy but might not be able to differentiate between different fault types. 

% The desire
In order to solve the detection and segmentation tasks within the same framework, fault diagnostic methods in open-set problems must find a compact description of the healthy class while correctly discriminating the fault types. If this can be achieved, fault detection and segmentation are possible with standard one-class and clustering methods. Since the performance of such methods in both tasks is mainly determined by the ability to obtain an informative feature representation of the CM, we focus on representation learning \cite{Bengio2013}.

% The obstacle
Methods using clustering algorithms perform increasingly better in a more compact and informative representation of the input space. An intuitive way to learn such a lower-dimensional representation (also referred to as embedding, latent, or feature space) is using autoencoders (AE). Without external induction, however, the latent representation of CM data typically shows an overlapping/entangled representation of the different fault types \cite{AriasChao2019}. As a result, such representation does not guarantee a more favorable distinction between fault types.

% The research question
Therefore, in this work, we investigate the hypothesis whether solely the knowledge of healthy and unlabeled conditions can regularize the latent representation of the CM data obtained with AE such that it separates unobserved fault types and improves fault detection. In other words, the research question is, does knowledge about the healthy conditions induce a more informative representation of the CM data that makes possible detection and segmentation of unlabeled and unobserved fault types?

% Paving the way to the method we want to propose
Variational Autoencoders (VAE) \cite{Kingma2013} provide more control over how the latent space is modeled. Therefore, they are generally preferred over traditional AEs for representation learning. Several extensions to VAEs have been proposed to encourage disentangled representations of the latent space (e.g. $\beta$-VAE \cite{Higgins2017}, Factor-VAE \cite{pmlr-v80-kim18b}, etc.) As already suggested in \cite{Locatello2019}, however, learning disentangled representations requires being explicit about the role of inductive biases and implicit supervision. Therefore, to answer the research question, we investigate simple but effective options to bring implicit supervision to the CM data's latent representation.

\begin{figure}[ht]
\centering
\includegraphics[width=8.7cm]{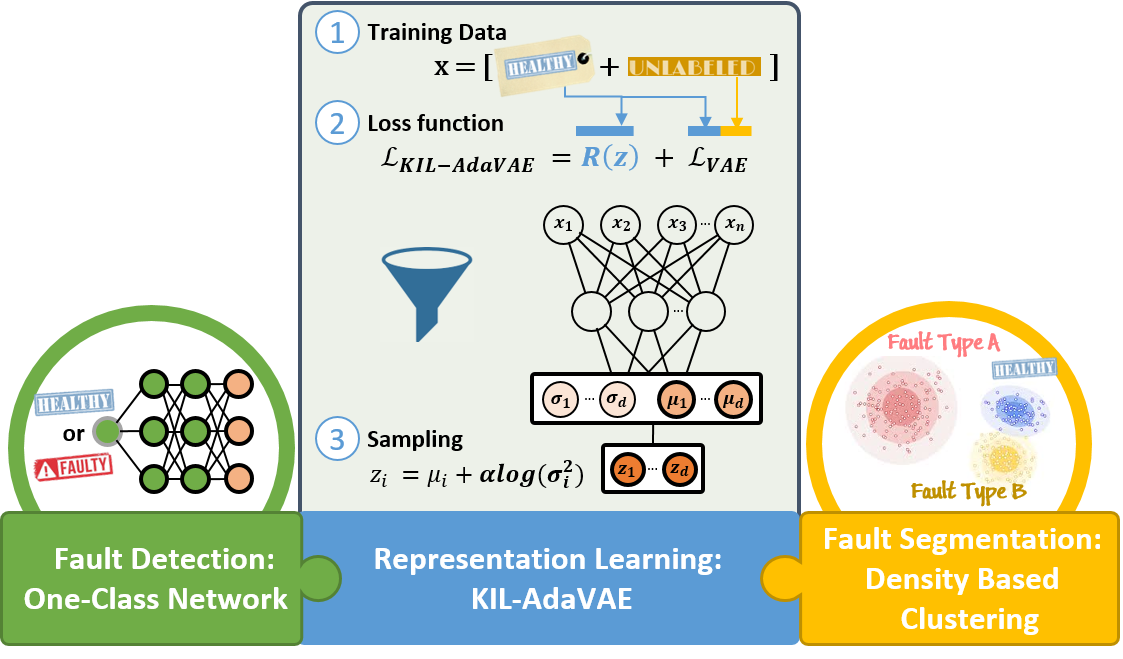} 
    \caption{Proposed solution strategy. \textbf{Representation learning} with three new features: 1) training a VAE with labelled and unlabeled samples, 2) an implicit supervision on the latent representation of the healthy conditions with an additional regularization term $R(z)$, and 3) a modified sampling procedure of the encoder network to enforce a latent space that is informative about the CM data. \textbf{Fault detection} uses a semi-supervised one class network. \textbf{Fault segmentation} uses a density based clustering.}
    \label{fig:idea}
\end{figure}

% The proposed method
In this work, we propose training a VAE with labeled and unlabeled samples and introduce two new features in the VAE design: implicit supervision on the latent representation of the healthy conditions, and implicit bias in the sampling process. Concretely, for implicit supervision, we propose to integrate the available knowledge on the healthy conditions in the VAE design through a new loss function. The use of labeled and unlabeled system conditions generates a more informative representation of the CM data in the latent space. The introduction of an additional loss term in the new loss function restricts the representation of healthy data and encourages a more distinctive representation of unobserved fault types. The modified sampling algorithm of the encoder network enforces a latent space that is more informative about the CM data. We refer to the proposed method as "knowledge induced variational autoencoder with adaptive sampling" (\text{KIL-AdaVAE}). The proposed method enables a compact and informative latent representation of the CM data that allows accurate segmentation, and detection of unseen fault types, using existing one-class and clustering algorithms. Figure \ref{fig:idea} shows an overview of the complete framework. The framework consists of three components: the proposed representation learning method (i.e., KIL-AdaVAE), the fault detection logic (a semi-supervised one-class neural network), and the fault segmentation method (a density-based clustering algorithm). %

% The experiments
The proposed framework is evaluated on fault detection and segmentation tasks in a new case study generated with the Advanced Geared Turbofan 30000 (AGTF30) dynamical model \cite{Chapman2017}. Real flight conditions and 17 different fault types are considered. In an extensive comparison, the proposed method is compared to other deep learning algorithms with different learning strategies, i.e., a) supervised one-class learning, b) supervised one-class learning with embedding, c) semi-supervised one-class learning with embedding. The results show that the proposed \text{KIL-AdaVAE} method is able to outperform others in terms of fault detection accuracy and fault segmentation capability of unlabeled fault types.

% The contribution
This paper presents two main contributions. From an application perspective, we formally introduce the problem of \textbf{open-set diagnostics} in complex industrial systems. For the case where the available dataset only consists of healthy labeled and unlabeled data, we propose a new representation learning method that, combined with existing one-class classification and clustering algorithms, results in a framework that outperforms other learning strategies in fault detection and segmentation. The dataset is openly available\footnote{https://www.research-collection.ethz.ch/handle/20.500.11850/430646} and can be used for comparison purposes in future work. From the methodological perspective, we propose KIL-AdaVAE, an extension of VAE for open-set diagnostics with three novelties: 1) labeled and unlabeled data are leveraged in the learning algorithm in a \textbf{semi-supervised} way, 2) \textbf{implicit supervision} on the latent representation of the healthy system conditions is incorporated, and 3) \textbf{implicit bias} is integrated in the sampling process of VAE in order to foster an informative latent representation of the CM data.

% ... And this is what is coming
The remainder of the paper is organized as follows. Section \ref{sec:related} outlines previous work on semi-supervised AD, AD and representation learning with autoencoders in the literature. In Section \ref{sec:Background}, the problem is formally introduced, and the background of the solution strategy is provided. In Section \ref{sec:proposed_method} the proposed method is described. In Section \ref{sec:Case_study}, the case study is introduced, and the experiment is explained. In section \ref{sec:Results} the results are given. Finally, a summary of the work and outlook are given in Section \ref{sec:Conclusion}.

\section{Related work}  \label{sec:related}
\textbf{Semi-supervised anomaly detection}. As already identified in \cite{Ruff2020Deep}, the term semi-supervised anomaly detection has been used to describe different AD settings. Most existing deep “semi-supervised” AD methods (such as \cite{Song2017AHS, Akay2018GANomalySA, Chalapathy2019, Borghesi2019}) only incorporate labeled normal samples in the learning algorithm. However, semi-supervised learning \cite{Chapelle} generally leverages unlabeled data in the main supervised task. In fact, SSL AD methods in \cite{Gornitz2013, Kiran2018, Min2018, Ruff2020Deep} also take advantage of the labeled and unlabeled data in the learning objective of the AD task. However, such a solution strategy involves a clustering assumption that contradicts the general scenario where anomalies (in our case, different fault types) are not necessarily similar to each another. The previously proposed solution strategy is, therefore, not suitable for unsupervised fault segmentation.

The combined use of labeled and unlabeled data can be limited to representation learning and not be part of the main supervised task objective. This corresponds to the M1 SSL model introduced by Kingma et al. in \cite{KingmaSemi}. Therefore, SSL anomaly detection methods can follow this solution strategy, which generally results in a two-step training process. First, labeled and unlabeled samples are used in a feature extraction task. In the subsequent step, the anomaly detector takes this representation as input but only uses labeled data for training. This AD solution strategy is also referred to as \textit{hybrid} in \cite{Ruff2020Deep} as it applies an anomaly detector to the latent space of an autoencoder. More precisely, it is an instance of Learning from Positive and Unlabeled Examples (LPUE) \cite{Bekker2020}. %Our proposed framework is based on this solution strategy. As main difference with SSL is that instead of leveraging the unlabeled data in the main supervised task (i.e., one-class classification), we induce implicit supervision in the unsupervised representation learning task with the available labeled data. 

\textbf{Anomaly detection with autoencoders.} As alternative to one-class classifiers, some researchers use the reconstruction error of an autoencoder as metric to detect anomalies, e.g., \cite{An2015VariationalAB, RIBEIRO201813, Borghesi2019}. This detection strategy is, however, similar to the one-class classifier strategy. The aim is to learn an AE model that accurately describes “healthy” system conditions. An anomaly is associated with any pattern that is poorly reconstructed by the AE. However, such a detection strategy is problematic in an open-set scenario. An AE trained with healthy labeled data and unlabeled data with potentially faulty conditions will result in low reconstruction error for similar faulty conditions. Therefore, reconstruction error might not be an adequate metric in open-set diagnostics.

\textbf{Representation learning with autoencoders.} The Infomax principle (\cite{Linsker1988}, \cite{Bell1995}, \cite{hjelm2018learning}) is the most commonly accepted explanation for representation learning in unsupervised settings. Under the Infomax principle, the autoencoder objective implicitly maximizes the mutual information between the input data ($x$) and the latent representation ($z$) (i.e., $I(x,z)$) under some regularization $R(z)$. In fact, regularized autoencoders are the predominant technique for representation learning. Hence, multiple variants have been used in diagnostics. Some examples are: contractive-autoencoders \cite{Zhang2020}, denoising autoencoders \cite{MENG2018448} or sparse autoencoders regularized with L1, Student-or Kullback-Liebler (KL) penalty \cite{SUN2016171, SHAO2018278, XU20181}. These methods use regularization on the representation $z$ to obtain statistical properties desired for some specific downstream tasks.  

A common desired statistical property of the latent representation is disentanglement. A disentangled representation $r(x)$ recovers the underlying factors of variation in observations $x$ such that each coordinate (i.e., $r(x)_i$ contains information about only one factor \cite{Locatello2019}. While this property is typically used for interpretability, predictive performance, and fairness reasons \cite{Locatello2020}, disentanglement can be beneficial for fault diagnostics. In particular, a disentangled representation shall improve the segmentation significantly. Several methods have been proposed to encourage disentangled representations in the latent space such as $\beta$-VAE \cite{Higgins2017}, Factor-VAE \cite{pmlr-v80-kim18b}, etc. However, Locatello et al. \cite{Locatello2019} demonstrated that the unsupervised learning of disentangled representations is theoretically impossible from i.i.d. observations without inductive biases.

\section{Background} \label{sec:Background}
In this section, we formally introduce the formulation of the open-set diagnostics problem. Also, we briefly introduce the basic concepts and notations related to one-class classification-based fault detection and Variational Autoencoders (VAE) as they are the building blocks of the framework and the method proposed in this work.

\subsection{Open-set diagnostics problem setup} \label{sec:Statement}
\begin{figure}[ht]
\centering
\includegraphics[width=8.7cm]{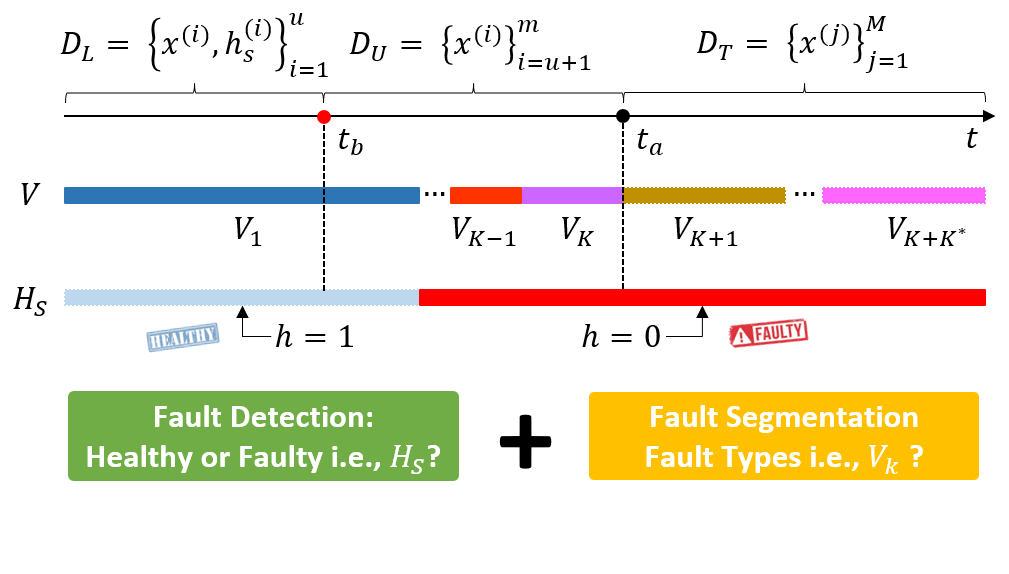} 
    \caption{Schematic representation of the problem. Training dataset $D$ has labeled ($D_L$) and unlabeled data ($D_U$). The test set ($D_T$) has only unlabeled samples. The true system conditions at any point in time are represented by the $H_S$ bar. Healthy conditions are represented in blue and faulty in red. $K$ fault types are present in $D_U$ and $K_*$ in $D_T$. The true system states are represented by the $V$ bar. Each fault type and the healthy conditions appear in a different color.}
    \label{fig:problem}
\end{figure}

We develop a diagnostics model at time $t_a$ from a multivariate time series of condition monitoring sensors readings $X = [x^{(1)}, \dots, x^{(m)}]^T$, where each observation $x^{(i)} \in R^{n}$ is a vector of $n$ raw measurements. The corresponding \textit{true} system health conditions (i.e., healthy or faulty) is partially known and denoted as $H_s = [h_s^{(1)}, \dots, h_s^{(m)}]^T$ with $h_s^{(i)} \in \{0,1\}$. Concretely, we consider the situation where certainty about healthy system conditions (i.e., $h_s^{(i)}=1$) is only available until a past point in time $t_b$ when the system condition was assessed and confirmed as healthy by maintenance engineers, e.g., during an inspection. The partial knowledge of the true health allows the definition of two subsets of the available data: a \textit{labeled} subset $\mathcal{D}_{L} = \{(x^{(i)}, h_s^{(i)})\}_{i=1}^{u}$ with $h_s^{(i)}=1$ corresponding to known healthy system conditions and an \emph{unlabeled} subset $\mathcal{D}_{U} = \{x^{(i)}\}_{i=u+1}^{m}$ with unknown system health conditions (i.e., the system health conditions are not associated with either healthy or faulty label). The unlabeled dataset is expected to contain data from both healthy and faulty system conditions. However, neither the number nor the type of faults is known. In this paper, we consider a scenario in which $K$ fault types are present in ${D}_{U}$. This represents the common and realistic situation where records about faults that have occurred in the field are not available at analysis time. A schematic representation of the problem setup is provided in Figure \ref{fig:problem}.

Given this set-up, the first task is to detect $K_{*}$ new faulty system conditions within $\mathcal{D_{T}} = \{x_{*}^{(j)}\}_{j=1}^{M}$ given the available dataset $\mathcal{D} = \mathcal{D}_{L} \cup {D}_{U}$ at time $t_a$. However, we also want to detect the faulty system conditions within ${D}_{U}$; which is a transductive learning problem. In other words, this first task involves determining a reliable direct or indirect mapping from the raw data ${X}$ to the possible system conditions (i.e., $\mathbf{\hat{h}_s}$) on $\{{D}_{U}, \mathcal{D_{T}}\}$. Hence, at testing time, the developed model is evaluated on fault types that were present in the unlabeled dataset and also on new (i.e., previously not observed) fault types. 

The second task is to provide an adequate unsupervised segmentation of fault types present in $\mathcal{D}_{U}$ and $\mathcal{D_{T}}$. We refer to $\textbf{V}=\{V_j|j=1, \dots, C\}$ as the partition of $\{{D}_{U}, \mathcal{D_{T}}\}$ according to the $C=K+K_{*}+1$ true system states, with $K+K_{*}$ faulty states and one healthy state.

\subsection{Fault detection} \label{sec:fd-methods}
The fault detection problem has been successfully addressed as a \textit{one-class classification} problem using hierarchical extreme learning machines (HELM) in \cite{Michau2017} and deep feed-forward neural networks in \cite{AriasChao2019}. Besides the different algorithms, these previous works share the same fault detection strategy, and in both cases, their detection performance outperforms the traditional one-class support vector machines \cite{Scholkopf2000}. Therefore, we resort to the same fault detection strategy that is briefly introduced in the following.

In one-class classification-based fault detection, the task turns to a regression problem that aims at discovering a functional map $\mathbf{\mathcal{G}}$ from the healthy system conditions to a target label $\mathbf{T} = \{h_s^{(i)} \; | \;  x^{(i)} \in S_{T}\}$ where $S_{T} \subsetneq \mathcal{D}_{L}$ is a training subset of $\mathcal{D}_{L}$. We refer to a neural network that discovers the functional map $\mathbf{\mathcal{G}}$ as the \textit{one-class network}. The output of the one-class network will deviate from the target value $\mathbf{T}$ when the inner relationship of a new data point $x^{(j)} \in \{\mathcal{D}_{U},\mathcal{D_{T}}\}$ does not correspond to the one observed in $S_{T}$. Therefore, the one-class network $\mathbf{\mathcal{G}}$ enables the definition of an unbounded similarity score $s_I(x^{(j)};\xi)$ of $x^{(j)}$ with respect to the healthy labeled data as follows:
\begin{equation} \label{eq:similarity}
  s_I(x^{(j)};\xi) =  \frac{\mid \mathbf{T}- \mathbf{\mathcal{G}}(x^{(j)}) \mid}{\xi}
\end{equation}
\begin{equation}
  \xi= P_{99.9}({\mid (\mathbf{T} - \mathbf{\mathcal{G}}(S_{V}) \mid}) 1.5
\end{equation}

where $\xi$ corresponds to a normalizing threshold. In this case study, we defined $\xi$ by the 99.9\% percentile of the absolute error of the prediction of $\mathbf{\mathcal{G}}$ in a validation set (i.e., $S_{V}$) extracted from $\mathcal{D}_{L}$ multiplied by a safety margin $1.5$. The percentile and the safety margin are model hyperparameters.

Hence, the fault detection logic is given by: 

\begin{equation}
    \hat{h}_{s}(x^{(j)}) = 
    \begin{cases} 
        1 & s_I(x^{(j)}; \xi) < 1  \\
        0 & \text{otherwise} 
    \end{cases}
    \label{eq:detection}
\end{equation}

Previous works considered embedding learning strategies combined with one-class classification to obtain the mapping function ${\mathcal{G}}$. The task has typically two parts. In a first step, a transformation $E: {X}_L \longmapsto \mathbf{z}_L$ of the input signals to a latent space $\mathbf{z_L} \in R^{\,m \times d}$ (with $d<n$) is found. The resulting latent space encodes optimal distinctive features of ${X_L}$ in an unsupervised way (i.e., without having information on the labels). In a second step, a functional mapping $\mathcal{G}_{sle}: E(X_{L}) \longmapsto \mathbf{T}$ from the latent space $\mathbf{z}_{L}=\{E(x^{(i)}) \; | \; x^{(i)} \in S_{T}\}$ to the target label $\mathbf{T}$ is learnt. This learning strategy is referred as \textit{supervised learning with embedding} (SLE).

Embedding learning strategies combined with one-class classification have also been proposed for anomaly detection tasks in \cite{chalapathy2018anomaly}. The main difference to the diagnostics approaches is the optimization objective of the one-class problem.

\subsection{Variational autoencoders} \label{sec:neuralnets}
VAEs are generative neural networks and assume that an observed variable $x$ is generated by some random process involving an unobserved random (i.e., latent) variable $z$. VAEs aim at sampling values of $z$ that are likely to have produced $x$ and compute $p(x)$ from those \cite{Doersch2016}. VAE models comprise an inference network (encoder) and a generative network (decoder). Contrarily to discriminative autoencoder models, the encoder and the decoder networks are probabilistic. The inference network $q_{\phi} (z \vert x)$ parametrizes the intractable posterior $p(z \vert x)$ and the generative network $p_{\theta} (x \vert z)$ parametrizes the likelihood $p(x \vert z)$ with parameters $\theta$ and $\phi$ respectively. These parameters are the weights and biases of a neural network. Typically, a simple prior distribution $p(z)$ is assumed (such as Gaussian or uniform).

The training objective of a generative model is to maximize the (marginal) likelihood of the data. This is,
\begin{align}
   \mathbb{E}_{p(x)}[\log p_{\theta}(x)] =  \mathbb{E}_{p(x)}[\mathbb{E}_{p(z)}[\log p_{\theta}(x \vert z)] 
\end{align}
However, direct optimization of the likelihood is intractable since $p_{\theta}(x) = \int_z p_{\theta}(x|z)p(z)dz$ requires integration \cite{Zhao2019}. Hence, VAEs consider an approximation of the marginal likelihood denoted Evidence Lower BOund (ELBO); which is a lower bound of the log likelihood (i.e., $\mathcal{L}_{\text{ELBO}} \leq \mathbb{E}_{p(x)}[\log p_{\theta}(x)]$)
\begin{align}
     \mathcal{L}_{\text{ELBO}} = \mathbb{E}_{p(x)}[\mathbb{E}_{q_{\phi}(x|z)}[\log p_{\theta}(x \vert z)] - D_{\text{KL}}(q_\phi(z \vert x) \vert \vert p(z))]
\end{align}
where $D_{\text{KL}}$ denotes the Kullback-Leibler (KL) divergence. Hence, the training objective of VAEs is to optimize the lower bound with respect to the variational parameters $ \phi $ and the generative parameters $ \theta $ 
\begin{align} \label{eq:ELBO}
   \max_{\phi, \theta} \mathbb{E}_{p(x)}[\mathbb{E}_{q_{\phi}(x|z)}[\log p_{\theta}(x \vert z)] - D_{\text{KL}}(q_\phi(z \vert x) \vert \vert p(z))]
\end{align}
The ELBO objective is the sum of two components. The first term is the reconstruction error and is equivalent to the training objective of an autoencoder. The second term, i.e., the KL divergence, is a distance measure between two probability distributions (i.e., $D_{KL}\geq 0$). Hence, the $D_{\text{KL}}$ term acts as a regularizer of $\phi$ trying to keep the approximate posterior $q_{\phi}(z|x)$ close to the prior $p(z)$.  

As a default assumption in VAE, the variational approximate posterior $q_{\phi}(z|x)$ follows a mutivariate Gaussian with diagonal covariance (i.e $q_{\phi}(z|x)= \mathcal{N}(z;\mu,\sigma^2 \mathbf{I})$) \cite{Kingma2013}. The distribution parameters of the approximate posterior $\mu$ and $\log \sigma^2$ are the non-linear embedding of the input $x$ provided by the encoder network with variational parameters $\phi$. The encoder output is, therefore, a parametrization of the approximate posterior distribution. With these assumptions, a valid local reparametrization of $z$ that allows to sample from the assumed Gaussian approximate posterior (i.e., $z^{(i)} \sim q_{\phi}(z|x^{(i)})$) is:
\begin{align}
   z^{(i)} = \mu^{(i)} + \sigma^{(i)} \odot\epsilon
\end{align}
with $\epsilon \sim \mathcal{N}(0,\mathbf{I})$.

There exists a variety of ways that extend the variational family to enrich the latent representation. One common way is to scale the $D_{KL}$ divergence term in the ELBO expression to encourage the desired properties of the resulting encoded representation. Higgins et al. \cite{Higgins2017} proposed the $\beta$-VAE that introduce a weight to the $D_{KL}$ term
\begin{align}
    \mathcal{L}_{\beta \text{-VAE}} = \mathbb{E}_{p(x)}[\mathbb{E}_{q_{\phi}(x|z)}[\log p_{\theta}(x \vert z)] - \beta D_{\text{KL}}(q_\phi(z \vert x) \vert \vert p(z))]
\end{align}
The $\beta$-VAE offers a trade-off between the information preservation, i.e., how well one can reconstruct $x$ from the $z$, and the capacity, i.e., how well the $z$ compresses information about $x$. Setting a $\beta>1$ forces the encoder to better match the factorized unit Gaussian (limiting the capacity of $z$) and deteriorates the reconstruction of $x$. The former can better be observed by reformulating the $D_{\text{KL}}$ term as follows:
\begin{align}
  \mathbb{E}_{p(x)}[D_{KL}(q_\theta(z \vert x) \vert \vert p(z))] = I(x;z) + D_{\text{KL}}(q(z)||p(z))
\end{align}
For $\beta \geq 1$ the $\beta$-VAE penalizes the mutual information between $z$ and $x$ (i.e., $I(x;z)$) but also enforces the so-called aggregated posterior $q(z)$ to factorize and match the prior $p(z)$. This factorization encourages a disentangled representation of the factors of variation in the data $x$. Hence, it can help to obtain an informative representation of independent factors of variations in the data.

\section{Proposed Method: Implicit supervision for open-set diagnostics} \label{sec:proposed_method}
\begin{figure*}[ht]
\centering
\includegraphics[width=16.5cm]{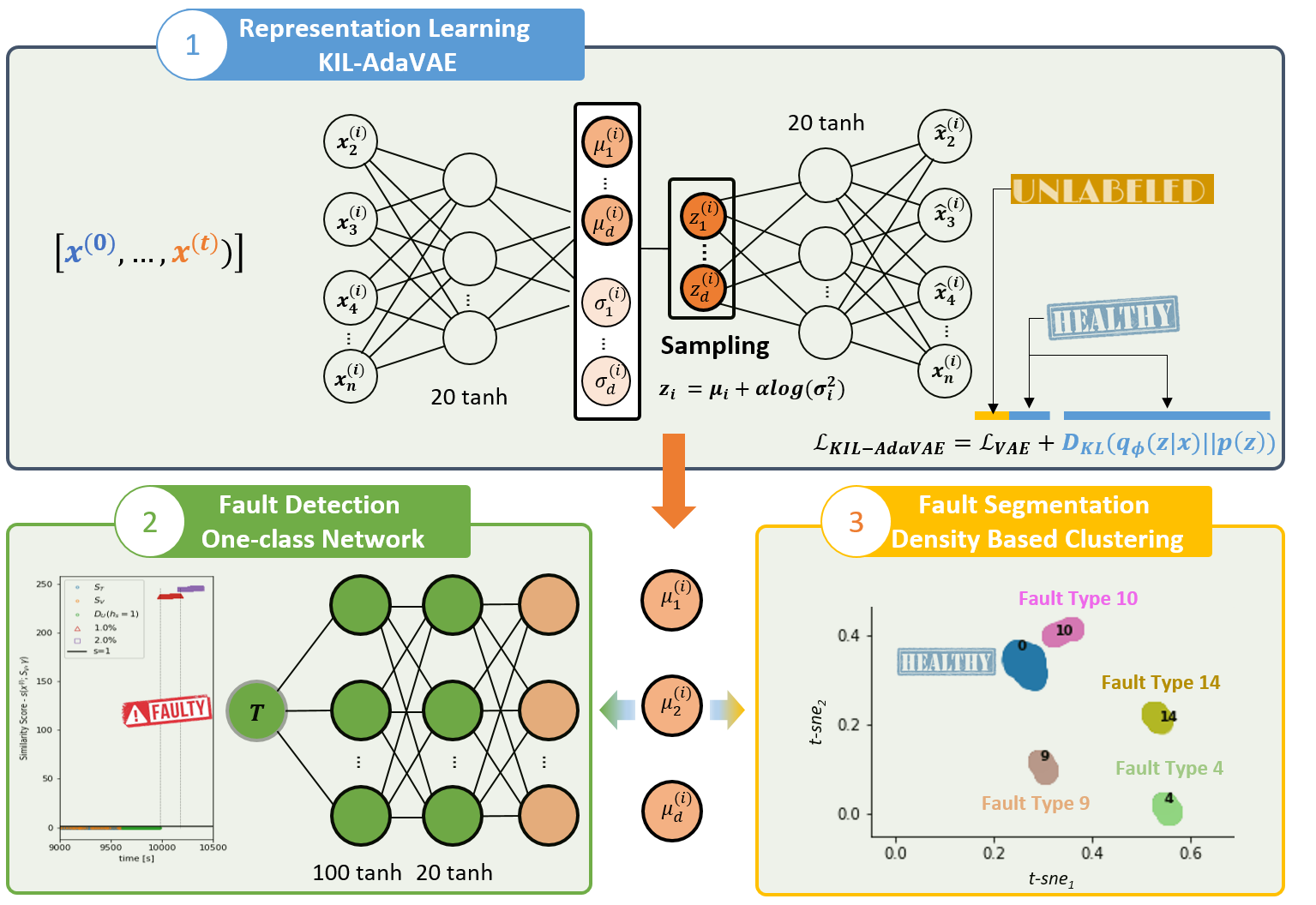} 
    \caption{Overview of the proposed framework. \textbf{Step 1 (representation learning)}: we train an autoencoder with labeled data with a known healthy condition and unlabeled data with an unknown health condition. KIL-AdaVAE also incorporates a new loss function and sampling algorithm.  \textbf{Step 2 (fault detection)}: taking as input the latent the representation of the CM data, we train a one-class network using only with healthy labeled data. The output of this network is used to define the fault detection logic. \textbf{Step 3 (fault segmentation)}: we apply a density-based clustering algorithm without any prior knowledge on the type and number of faults to the latent representation of the CM data for fault segmentation.}
    \label{fig:framework}
\end{figure*}

In the following, we introduce the proposed method and the complete framework for open-set diagnostics (see Figure \ref{fig:framework}). First, we focus on fault detection and explain the generalization of the one-class classification strategy in \cite{AriasChao2019} to the semi-supervised setting. We then introduce the proposed KIL-AdaVAE method for representation learning. Finally, we explain the clustering method used for fault segmentation.

\subsection{Fault detection}
\textbf{Semi-supervised one-class learning}. We build on the previous work \cite{AriasChao2019} to generalize the supervised learning with an embedding strategy (SLE) to a semi-supervised setting. As discussed earlier, SLE is designed to use exclusively healthy labeled data in the learning algorithm. This means that when a SLE strategy is used in an open-set problem, the \textit{unlabeled} dataset $\mathcal{D}_{U}$ is not used at training time. However, the discovery of the mapping function, $\mathcal{G}$, can also be formulated as a semi-supervised learning problem to take advantage of the relevant information that may be present in $\mathcal{D}_{U}$. Concretely, semi-supervised learning methods consider both labeled and unlabeled data during training. Therefore, we define an alternative unsupervised problem $F:X \longmapsto \mathbf{z} \longmapsto X$ to include all the available sensor data $X$ and obtain a latent representation $\mathbf{z} \in R^{\,n \times d}$ that encodes features of healthy and potentially faulty states. Similarly as before, in a second step, we find a supervised mapping $\mathcal{G}_{ssl_{M1}}: \mathbf{z}_{L} \longmapsto \mathbf{T}$. This framework is similar to the M1 model introduced by Kingma et al. in \cite{KingmaSemi}.

\subsection{Representation learning}
Regularization over the latent space is a natural strategy in representation learning. In particular, the $D_{\text{KL}}$ encourages the encoded representation of the training data to match the factorized unit Gaussian. Carefully considering this property, we can create a discriminative representation of the healthy system conditions with respect to the faulty ones. %Such a representation is, therefore, very informative for fault detection and segmentation tasks.

\textbf{Knowledge induced learning with adaptive sampling variational autoencoder (\text{KIL-AdaVAE}).} In a SSL setting, a mix of healthy and faulty conditions might be present in the training data. Therefore, forcing the encoded representation of the training data to match the factorized unit Gaussian does not necessarily promote a discriminative representation of the healthy labeled system conditions. Instead, to achieve this, we propose introducing an implicit supervision on the latent representation of the healthy labeled system conditions (i.e., $S_{T}$). Hence, we incorporate an additional loss-term to the ELBO and optimize the following loss:
\begin{align}
   \mathcal{L}_{\text{KIL-AdaVAE}} &= {\mathcal{L}_{\text{ELBO}}} - \gamma D_{KL}(q_\phi(z \vert x) \vert \vert p(z))_{S_T}
\end{align}

The additional loss-term forces the representation of the \textbf{healthy data} (i.e., $S_{T}$) to match the factorized unit Gaussian. With this new loss-term we bring implicit supervision to the unsupervised learning task of VAE. It is worth pointing out that, this regularization objective is different from the one in \cite{Ruff2020Deep}. In that case, the supervision is explicit and the regularization objective includes the healthy label $T$, i.e., $R(z, T)$. The weight of this new divergence term relative to the standard $D_{\text{KL}}$ in the ELBO is controlled by the hyperparameter $\gamma$. We refer to this learning strategy as \textit{knowledge induced learning} since it integrates the knowledge on the healthy system condition in the training process.

\textit{Adaptive Sampling.} The ELBO objective imposes a balance between reconstruction and regularization. This balance reduces the information about the input in the latent representation. Therefore, to foster a latent representation that also preserves the information about the input, we propose the following simple but effective modification of the sampling approach:
\begin{align}
  z^{(i)} & = \underbrace{\mu^{(i)}}_{\text{location}} + \underbrace{\alpha \log ((\sigma^{(i)})^2)}_{\text{scale}} \odot\epsilon
\end{align}
The proposed sampling approach replaces the standard scale term i.e., $\sigma^{(i)}$ with an approximation of the entropy of an isotropic Gaussian i.e., $\alpha \log ((\sigma^{(i)})^2)$. This approximation holds because, under the VAE framework, the samples from the posterior $z^{(i)} \sim q_{\phi}(z|x^{(i)})$ are assumed to follow an isotropic Gaussian, $z_k \sim N(\mu_k, \sigma_k^{2} \mathbf{I})$ with $\sigma_k >0$, for which the entropy has the following close form \cite{Ahmed1989}:  
\begin{multline}
  \mathcal{H}(z^{(i)}) = \frac{d}{2}  + \frac{d}{2}\log(2\pi) + \frac{d}{2}\log ((\sigma^{(i)})^2) \propto \log ((\sigma^{(i)})^2)
\end{multline}

This functional \textit{scale} sampling imposes an implicit bias to the standards VAE sampling to reduce its variance when $z \sim N(0, \mathbf{I})$. Therefore, it encourages informative samples $z^{(i)}$ while keeping the distribution parameters of the approximate posterior $\mu^{(i)}$ and $\sigma^{(i)}$ close to the prior. This effect is best reasoned with a graphical example.

\begin{figure}[ht] 
\centering
\includegraphics[width=8.5cm]{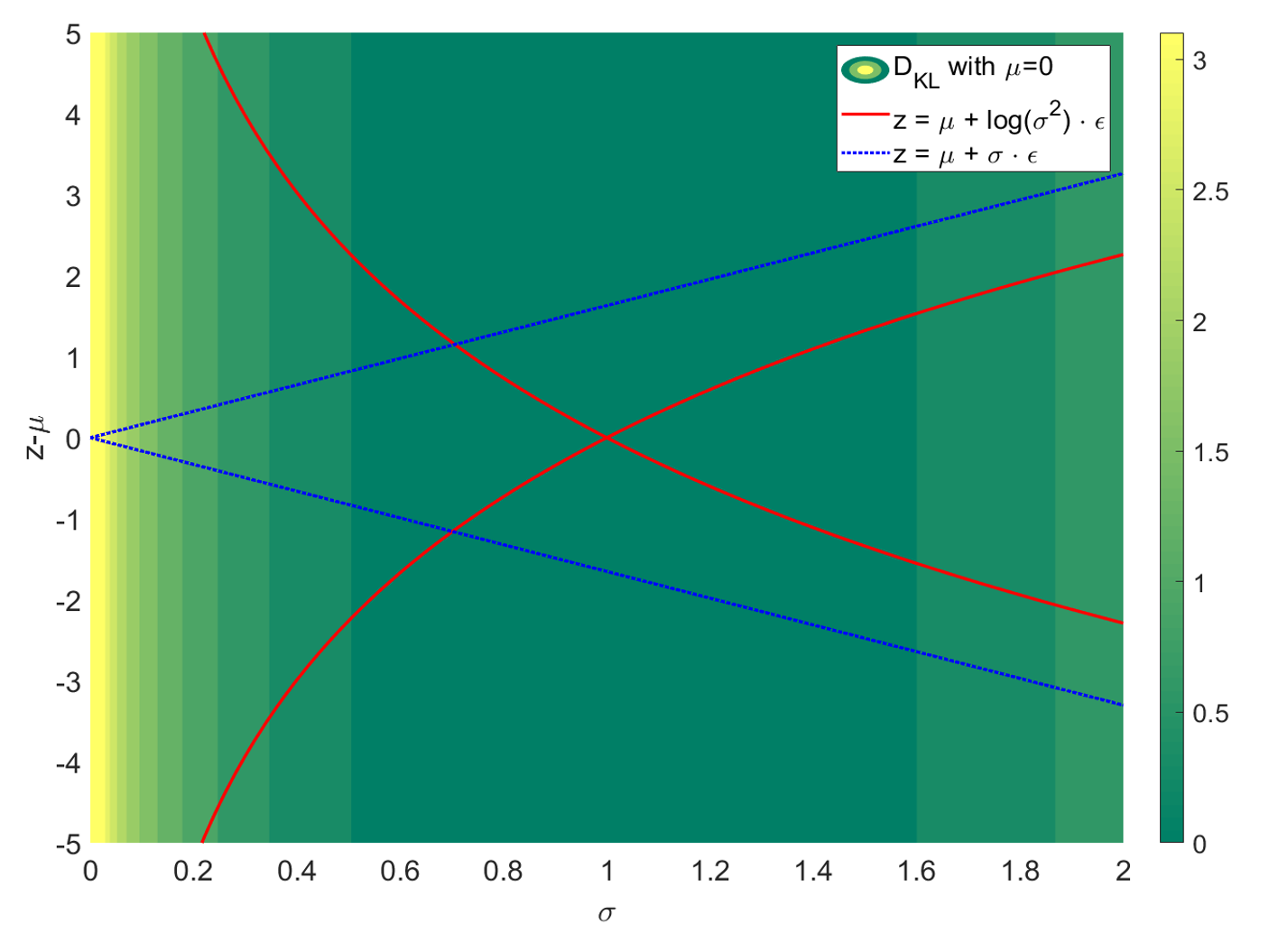}
\caption{95\% (top) and 5\% (bottom) percentiles of the $z-\mu$ term for standard VAE sampling (blue dashed lines) and the proposed adaptive sampling (red continuous lines). The region between these percentiles represent ($z-\mu$) is the possible sampling region of each method as function of $\sigma$. The contour plot shows the Kullback-Leibler divergence (i.e., $D_{KL}$) is a function $\sigma$ for $\mu=0$. The $D_{KL}$ term regularize the reconstruction error by penalizing values of $\mu \neq0$ and $\sigma \neq1$. $D_{KL}$ has a minimum at $\mu=0$ and $\sigma=1$.}
\label{fig:sampling}
\end{figure}

The impact of the functional \textit{scale} in the sampled $z^{(i)}$ is given by the term $z^{(i)} - \mu^{(i)}$. Figure \ref{fig:sampling} provides a comparison of the $z^{(i)} - \mu^{(i)}$ terms as a function on $\sigma^{(i)}$ for the proposed sampling (i.e., $z = \mu + log(\sigma^2) \cdot \epsilon$) and the standard VAE sampling (i.e., $z = \mu + \sigma \cdot \epsilon$). The 95\% and 5\% percentiles of $z^{(i)} - \mu^{(i)}$ with both sampling types are depicted. The contour plot in the background shows the values of $D_{\text{KL}}$ as a function of $\sigma$ for $\mu=0$. Kullback-Leibler divergence is a function of $\mu$ and $\sigma$ and has a minimum at $\mu=0$ and $\sigma=1$. We can observe that the proposed functional scale results in samples $z^{(i)}$ with low variance when $D_{KL} \approx 0$. Concretely, if $\sigma^{(i)} \approx 1$ (i.e., $\log (\sigma^{(i)})^2 \approx 0$), the generated samples $z^{(i)}$ are maintained close to the mean $\mu^{(i)}$ (i.e. $z^{(i)} \approx \mu^{(i)}$). Since $\sigma^{(i)}$ is matching the prior, the $\text{ELBO}$ objective encourages values of $\mu^{(i)}$ that are informative (i.e., low reconstruction error) while being close to $\mu^{(i)} \approx 0$. In contrast, a standard sampling leads to samples $z^{(i)}$ with higher variance in areas of low $D_{KL}$; which results in a loss of information about $x^{(i)}$ in the encoded representation $\mu^{(i)}$ and the sample $z^{(i)}$. Besides this descriptive reasoning, experimental evidence demonstrates the impact of the sampling on the reconstruction error (i.e., $\mathcal{L}_{\ell_2}$), the optimisation objective (i.e., $\mathcal{L}_{\beta\text{-VAE}}$) and the mutual information between $x^{(i)}$, $\mu^{(i)}$ and $z^{(i)}$ and is detailed in Section \ref{sec:betaVAE}.

In contrast to other representation learning methods aiming to control the mutual information between the input and the latent variable (e.g., \cite{Alemi2018FixingAB, hjelm2018learning, Zhao2019}), this simple technique does not require the modification of the learning objective.

The proposed sampling deviates from an isotropic Gaussian. However, for simplicity, we regularize the optimisation objective with a $D_{\text{KL}}(q_\phi(z \vert x)|p_{\theta}(z))$ computed under the hypothesis that both $p_{\theta}(z)$ and $q_{\phi}(z|x)$ are Gaussian distributions. In this way, the $D_{\text{KL}}$ term can be computed analytically as follows:
\begin{multline}
      -D_{KL}(q_\phi(z \vert x) \vert \vert p(z)) = \\ 
      \sum_{j=1}^d (1-\log (\sigma_j^{(i)})^2 - (\mu_j^{(i)})^2 -(\sigma_j^{(i)})^2\big)
\end{multline}

The \text{KIL-AdaVAE} model implements this solution strategy with $\alpha=d/2$ and $\gamma=n$. The entire proposed procedure is provided in Algorithm \ref{al:kil-adavae}.

\vspace{-0mm}
\begin{algorithm}[ht]
\caption{Fault detection with knowledge induced learning with adaptive sampling VAEs.} \label{al:kil-adavae}
%\SetAlgoLined
%\DontPrintSemicolon

\textbf{Input}: $\{x^{(i)}\}_{i=1}^m \in \{S_T, D_U\}$ Labeled and unlabeled samples\\
$\theta, \phi \gets \text{Initialize parameters VAE}$\\
$X_L=Upsample(\{x^{(i)}\}_{i=1}^u) \; \text{Upsample $S_T$ to size $m$}$ with replacement\\
% Unsupervised VAE
\While{$i \leq E$}{
$g \gets \nabla_{\theta, \phi} \mathcal{L}_{\text{KIL-AdaVAE}}(x, x_{L})$\\
$\theta, \phi \gets \text{Update parameters using gradient g}$ \\
}
% Supervised - One Class training
\textbf{Input}: $\{x^{(i)}, h_s^{(i)}\}_{i=1}^u \in  S_T$ Labeled samples and their labels
\For{$i \in \{1,\dots,u\}$}{
$\mu^{(i)} \sim q_{\phi}(z|x^{(i)})$\\
}
$\xi \gets MaxMin(\mu_{S_T})$\\

$\mathcal{H} \gets \text{Initialize parameters One-Class Network}$\\
\While{$i \leq E_s$}{
$g \gets \nabla_{\mathcal{H}} \mathcal{L}_{\text{One-Class}}$\\
$\mathcal{H} \gets \text{Update parameters using gradient g}$ \\
}
% Detection
\textbf{Input}: $\{x^{(j)}\}_{i=u+1}^m \in D_U$ Unlabeled samples\\
\For{$i \in \{1,\dots,m\}$}{
$\mu^{(i)} \sim q_{\phi}(z|x^{(i)})$\\
$\mu_N^{(i)} \gets MaxMin(\mu^{(i)}, \xi)$\\
$s_k(\mu^{(i)};S_V) = \frac{\mid 1- \mathbf{\mathcal{G}}(\mu^{(i)}) \mid}{ P_{99.99}({\mid (1 - \mathbf{\mathcal{G}}(S_V) \mid})}$\\
$\hat{h}_{s}(x^{(i)})  \gets \text{from Equation } \ref{eq:detection}$
}
\textbf{Input}: $\{x^{(j)}\}_{j=1}^M \in D_T$ Test samples\\
\For{$j \in \{1,\dots,M\}$}{
$\mu^{(j)} \sim q_{\phi}(z|x^{(j)})$\\
$\mu_N^{(j)} \gets MaxMin(\mu^{(j)}, \xi)$\\
$s_k(\mu^{(j)};S_V) = \frac{\mid 1- \mathbf{\mathcal{G}}(\mu^{(j)}) \mid}{ P_{99.99}({\mid (1 - \mathbf{\mathcal{G}}(S_V) \mid})}$\\
$\hat{h}_{s}(x^{(j)})  \gets \text{from Equation } \ref{eq:detection}$
}
\end{algorithm}

\subsection{Fault segmentation} \label{sec:fc-methods}
The latent space $z$ obtained with semi-supervised and supervised with embedding learning methods provide an alternative representation of the input signals. This encoded representation can reveal hidden patterns that make faulty system conditions clearly detectable. Hence, we applied a density-based clustering algorithm to the latent space $z$ to discover unknown system conditions present in the dataset $\{D_U, D_T\}$. Here, we performed fault segmentation on both $\{D_U\}$ and  $\{D_T\}$ since the fault classes of $\{D_U\}$ are also unknown. Concretely, we used the density-based algorithm \textit{Ordering points to identify the clustering structure} (i.e., OPTICS) to group points $z^{(i)} \in \{D_U, D_T\}$ that are close to each other based on a metric of distance (i.e., Euclidean distance) and a minimum number of points. We selected the OPTICS algorithm because of its ability to detect meaningful clusters in data with varying density. Hence, OPTICS addresses a major weakness of the commonly applied \textit{density-based spatial clustering of applications with noise} (DBSCAN) algorithm \cite{Ester1996}.

Visualization of the resulting clusters is carried out with a two-dimensional representation of the high dimensional latent space ($z$) using the t-Distributed Stochastic Neighbor Embedding (t-SNE) algorithm \cite{VanDerMaaten2008}.

\section{Case study} \label{sec:Case_study}

\subsection{The AGTF30 dataset} 
\begin{figure}[ht]
\centering
\includegraphics[width=8.7cm]{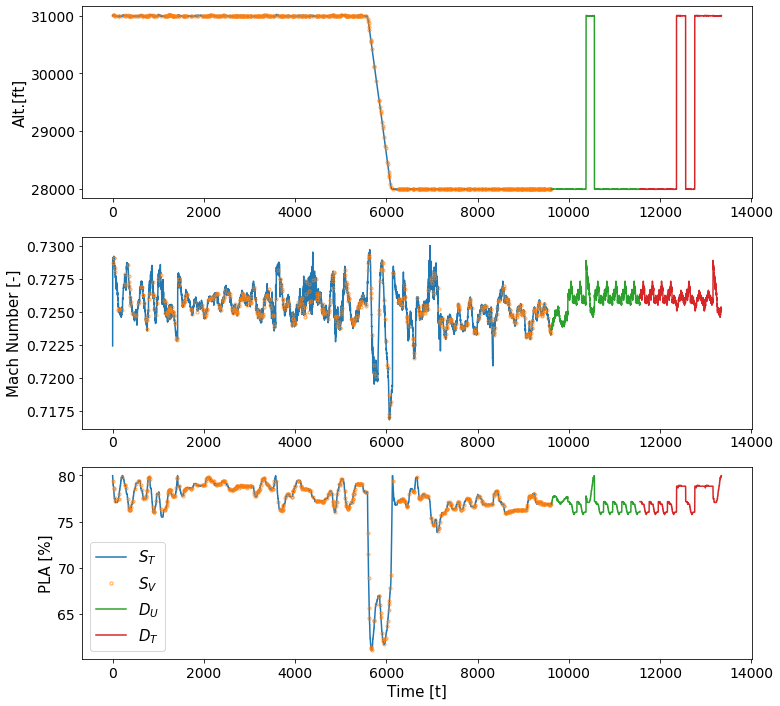}
    \caption{Flight envelope given by the traces of altitude, flight Mach number and power lever angle (PLA). Four datasets are shown: $S_T$ (blue), $S_V$ (orange), $D_U$ (green) and $D_T$ (red).}
    \label{fig:operation_profile}
\end{figure}

A new dataset was designed to evaluate the proposed method. The AGTF30 dataset provides simulated CM data of an advanced gas turbine during flight. The dataset was synthetically generated with the AGTF30 (Advanced Geared Turbofan 30k lbf) dynamical model \cite{Chapman2017}. Real flight conditions recorded on board of a commercial jet \cite{DASHlink} were taken as input to the AGTF30 model. Figure \ref{fig:operation_profile} shows the corresponding flight envelope given by the traces of altitude ($Alt$), flight Mach number ($MN$) and power lever angle ($PLA$). Two distinctive stable flying altitudes and a rapid transient maneuver for altitude adaptation can be observed. The labeled dataset $\mathcal{D}_L$ (blue) consists of multivariate steady-state responses (i.e., $x^{(i)} \in R^{13}$) of the AGTF30 model during 10.000 s of flight at cruise with a healthy system condition (i.e., $h_s=1$). The unlabeled (green) and test (red) datasets $\{\mathcal{D}_U, \mathcal{D_{T}}\}$ contain $C=17$ concatenated time series of model responses resulting from faulty engine conditions. Table \ref{tb:faults} shows a detailed overview of the eight faults present in the unlabeled and the nine faults types in the test dataset. The test set contains fault types affecting components not present in $\mathcal{D}_U$ (e.g., C=3, 4 \& 17). Each fault corresponds to an individual component fault and has a duration of approx. 200 s. The flight conditions in which faults were induced are randomly assigned from a subset of three operation intervals extracted from the cruise envelope (see Table \ref{tb:flight_conditions}). The unlabeled dataset also includes data from 500 s of operation when the engine is healthy. The unlabeled and test datasets $\{\mathcal{D}_U, \mathcal{D_{T}}\}$ are, therefore, a set of $C+1$ truncated system conditions. It is worth pointing out that the combination of flight conditions and fault types (i.e., component affected and fault magnitude) increases the differences between $\mathcal{D}_U$ and $\mathcal{D}_T$ and adds difficulty to the diagnostics task. No additional noise was added to the model response since the input values are already noisy. The sampling frequency of the simulation is 1Hz.

%%%%%%%%%%%%%%% begin table - Comp  %%%%%%%%%%%%%%%%%%%%%%%%%%
\begin{table}[ht]
\caption[Table tb:faults]{Overview of Generated Faults}
\begin{center}
    \label{tb:faults}
\begin{tabular}{clccc}
\hline
C              & Component       &  Magnitude    &  Flight        & Dataset          \\
\hline
1              & Fan Efficiency  &   1.0 \%      &  $S_1$         & $\mathcal{D}_U$   \\ 
2              & Fan Efficiency  &   1.5 \%      &  $S_1$         & $\mathcal{D_{T}}$ \\ 
3              & Fan Capacity    &   1.0 \%      &  $S_1$         & $\mathcal{D_{T}}$ \\
4              & LPC Efficiency  &   1.5 \%      &  $S_1$         & $\mathcal{D_{T}}$ \\
5              & LPC Capacity    &   1.5 \%      &  $S_1$         & $\mathcal{D_{T}}$ \\ 
6              & LPC Capacity    &   2.0 \%      &  $S_1$         & $\mathcal{D}_U$   \\
7              & HPC Efficiency  &   1.0 \%      &  $S_2$         & $\mathcal{D}_U$   \\ 
8              & HPC Efficiency  &   1.5 \%      &  $S_3$         & $\mathcal{D_{T}}$ \\ 
9              & HPC Efficiency  &   2.0 \%      &  $S_1$         & $\mathcal{D}_U$   \\
10             & HPC Capacity    &   1.0 \%      &  $S_1$         & $\mathcal{D_{T}}$ \\ 
11             & HPC Capacity    &   2.0 \%      &  $S_2$         & $\mathcal{D}_U$   \\ 
12             & HPT Efficiency  &   1.0 \%      &  $S_1$         & $\mathcal{D}_U$   \\
13             & HPT Efficiency  &   1.5 \%      &  $S_3$         & $\mathcal{D_{T}}$ \\
14             & HPT Efficiency  &   2.0 \%      &  $S_1$         & $\mathcal{D}_U$   \\
15             & HPT Capacity    &   1.5 \%      &  $S_3$         & $\mathcal{D_{T}}$ \\ 
16             & HPT Capacity    &   2.0 \%      &  $S_1$         & $\mathcal{D}_U$   \\
17             & LPT Capacity    &   1.0 \%      &  $S_2$         & $\mathcal{D_{T}}$ \\
\hline
\end{tabular}
  \end{center}
\end{table}

%%%%%%%%%%%%%%% begin table - Comp  %%%%%%%%%%%%%%%%%%%%%%%%%%
\begin{table}[ht]
\caption[Table tb:flight conditions]{Overview of Flight Conditions During Faults}
\begin{center}
    \label{tb:flight_conditions}
\begin{tabular}{clcc}
\hline
Symbol   & Alt [ft]        &  Match [-]     &  PLA[\%]      \\
\hline
$S_1$    & 28.0k           &   0.727-0.726  &  77.2-75.8    \\ 
$S_2$    & 31.0k - 29.6k   &   0.729-0.722  &  80.0-66.9    \\ 
$S_3$    & 31.0k - 30.0k   &   0.727-0.725  &  78.9-78.7    \\
\hline
\end{tabular}
  \end{center}
\end{table}

\subsection{Pre-processing}
The input space $X$ to the neural network models is normalized to a range $[-1, 1]$ by min/max-normalization. A validation set $S_{T}$ comprising 6 \% of the labeled data for all the models was chosen for setting the hyperparameters.

Table \ref{tb:cm_signals} provides a detailed overview of the model variables included in the CM signals. The variable name corresponds to the internal variable name used in \cite{Chapman2017}. The descriptions and units are reported as provided in the model documentation.

%%%%%%%%%%%%%%% begin table - [W,X_s]  %%%%%%%%%%%%%%%%%%%%%%%%%%
\begin{table}[ht]
\caption[Table tb:CM]{Condition monitoring signals}
\begin{center}
    \label{tb:cm_signals}
\begin{tabular}{cllc}
\hline
\#           & Symbol       &  Description                       & Units          \\ \hline
1             & alt         & Altitude                           & ft             \\
2             & MN          & Flight Mach number                 & -              \\
3             & PLA         & Power Lever Angle                  & \%             \\
4             & Wf          & Fuel flow                          & pps            \\
5             & N\_LPC      & Low pressure (LP) shaft speed      & rpm            \\
6             & N\_HPC      & High pressure shaft (HP) speed     & rpm            \\
7             & S45\_Tt     & Total temperature at HPT outlet    & $^{\circ}$R    \\
8             & Pa          & Ambient static pressure            & psi            \\
9             & S2\_Pt      & Total pressure at fan inlet        & psi            \\
10            & S25\_Pt     & Total pressure at LPC outlet       & psi            \\
11            & S36\_Pt     & Total pressure at HPC outlet       & psi            \\
12            & S5\_Pt      & Total pressure at LPT outlet       & psi            \\
13            & VAFN        & Variable area fan nozzle           & $in^2$         \\ \hline
\end{tabular}
 \end{center}
\end{table}

\subsection{Baseline models and variants}
In an extensive evaluation of the proposed \text{KIL-AdaVAE} approach, we compared its performance to nine alternative models divided into four learning strategies:

\textbf{Supervised one-class learning with embedding}. As the supervised learning with embedding (SLE) is our baseline learning strategy, we implemented previous discriminative models based on hierarchical extreme learning machines (\text{SLE-HELM}) \cite{Michau2017} and deep feed-forward autoencoders (\text{SLE-AE}) \cite{AriasChao2019}. We also considered three feed-forward Variational Autoencoder models: \text{SLE-VAE}, \text{SLE-$\beta$-VAE}, and \text{SLE-AdaVAE}. To make the comparison more challenging, we deliberately provided the \text{SLE-$\beta$-VAE} model an unfair advantage by selecting the $\beta$ hyperparameter to maximize the fault detection accuracy (i.e., $\beta=5$). The \text{SLE-AdaVAE} model is a variant of the \text{SLE-$\beta$-VAE} model that includes the proposed adaptive sampling with $\alpha=d/2$.

\textbf{Semi-supervised one-class learning}. For semi-supervised learning, we considered two models based on feed-forward Variational Autoencoders: \text{SSL-M1-VAE} and \text{SSL-M1-AdaVAE}. \text{SSL-M1-VAE} model is a standard VAE trained with labeled and unlabeled samples. \text{SSL-M1-AdaVAE} also incorporates the proposed adaptive sampling with $\alpha=d/2$. Therefore, these two models incorporate individually two features of the proposed \text{KIL-AdaVAE} model. 

\textbf{Implicitly-supervised one-class learning}. We also considered a VAE model that only incorporates the implicit supervision of the latent representation of labeled data (\text{KIL-VAE}). For a clear comparison, the \text{KIL-VAE} model uses $\gamma=n$.

\textbf{Supervised one-class learning}. To complete the learning spectrum, we also considered a supervised learning (SL) strategy (i.e., a direct functional mapping $\mathcal{G}_{sl}: {X}_{L} \longmapsto \mathbf{T}$). This learning strategy has also been proposed in \cite {pmlr-v80-ruff18a} for anomaly detection. The \text{SL-FF} model implements a feed-forward one-class neural network.

\subsection{Network architectures} \label{sec:networks}
The learning strategies described above require two overall network architectures. On one hand, \emph{semi-supervised} and \emph{supervised with embedding} learning strategies require an autoencoder network in addition to the \textit{one-class network}. On the other hand, the supervised learning strategy requires only a \textit{one-class network}. 

To evaluate the different models, we separate the effect of regularization in the form of model and learning strategies from other inductive bias in the form of choice of neural network architectures. Therefore, we define a shared one-class network and AE architectures for semi-supervised and supervised with embedding models. Concretely, we use the same network architectures as in \cite{AriasChao2019}, which is a similar diagnostics problem. The network architectures are briefly described in the following.

\textbf{One-class network}. The one-class network architecture uses three fully connected layers. The first hidden layer has 20 neurons, and the last hidden layer has 100 neurons. The network ends with a linear output neuron. Hence, in a compact notation, we refer to the \textit{one-class network} architecture as $[d, 20, 100, 1]$. The \textit{tanh} activation function is used throughout the network. 

\textbf{Autoencoder networks}. The autoencoder models use the same encoder architecture with one hidden layer of size 20 neurons and a latent space of 8 neurons (i.e., $d=8$). In compact notation, we refer to the \textit{autoencoder network} architecture as $[13, 20, 8, 20, 13]$. The same architecture is also implemented for the \textit{SLE-HELM} model.

The \text{supervised learning} strategy requires only one network. Hence, to enable a clear comparison to the representation learning capabilities of supervised with embedding and semi-supervised models, the structure of the \textit{SL-FF} has five layers with the following architecture $[13, 20, 8, 20, 100, 1]$. The first part of the network, from the input to the second hidden layer, follows the architecture of the \textit{encoder network}. The remaining of the network is the same as the one-class in the SLE and SSL. Therefore, we unconventionally refer to the first part of the network as the \textit{equivalent encoder network} and the second hidden layer as the supervised latent space $z_s$. 

\subsection{Training set-up}
The optimization of the network weights for each of the models except HELM was carried out with mini-batch stochastic gradient descent (SGD) and with the \textit{Adam} algorithm \cite{Kingma2014Adam}. HELM used the training method described in \cite{Michau2017}. \textit{Xavier} initializer \cite{Glorot} was used for the weight initializations. The learning rate was set according to Table \ref{tb:Settings}. The batch size for the autoencoders configurations was set to 512 and to 16 for the one-class classifiers. Similarly, the number of epochs for autoencoder training was set to 800 and for the one-class classifiers to 300.

%%%%%%%%%%%%%%% begin table   %%%%%%%%%%%%%%%%%%%%%%%%%%
\begin{table}[ht]
  \caption[Table caption text]{Default training parameters}
  \begin{center}
    \label{tb:Settings}
    \begin{tabular}{lcccc}
      \hline
      Model             & LR             &  Batch size   &    Epochs    \\ \hline
       SL               & 0.001          &      16       &     300     \\
       SLE, SSL \& KIL  & 0.005/0.001    &    512/16     &   800/300   \\ 
      \hline
    \end{tabular}
  \end{center}
  \vspace{-5mm}
\end{table}
%%%%%%%%%%%%%%%% end table %%%%%%%%%%%%%%%%%%%%%%%%%%%%%

\subsection{Evaluation metrics}
The performance of the proposed method was evaluated and compared to alternative neural network models on the selected fault diagnostics task: detection of unknown faults (i.e., estimation of $h_s$), and unsupervised segmentation of the fault types (i.e., estimation of $\mathbf{V}$). We also evaluated the capability of different models to find a useful transformation that is informative of the system condition (i.e., representation learning). For each of the three groups, we considered targeted evaluation metrics that are defined in the following.

\textbf{Metrics for fault detection} Given the combined dataset $\{D_{U}, D_{T}\}$ with the true health state $h_s$ and the corresponding estimated health state $\hat{h}_s$, we evaluated the detection performance of the different neural network models as the accuracy of a binary classification problem. Three metrics for binary classification were considered: accuracy ($\text{Acc}$), true positive rate ($\text{TP}$), false positive rate ($\text{FN}$).

\textbf{Metrics for fault segmentation} Since the discovery and classification of fault types from unlabeled data ($\{D_{U}, D_{T}\}$) were addressed as a clustering problem, we used the following four standard clustering metrics to evaluate the fault segmentation performance \cite{XuanVinh2010}: number of clusters ($C$), homogeneity ($h$), completeness ($c$), and Adjusted Mutual Information ($AMI$). We selected these metrics because of their strong mathematical foundation since they are rooted in information theory, and their ability to analyze non-linear similarities.

\textbf{Metrics for representation learning} To provide a deeper understanding of the different model capabilities and to explain their diagnostic performance, we defined three auxiliary metrics:

\textit{Adjusted Mutual Information Gain ($\text{AMIG}$)}: The adjusted mutual information ($\text{AMI}$) evaluates the information that an alternative cluster representation of the data $\textbf{U}=\{U_i|i=1, \dots, R\}$ shares with the true segmentation of the system conditions $\textbf{V}=\{V_j|j=1, \dots, C\}$. $\text{AMI}$ takes the value of 1 when the two partitions $U$ and $V$ are identical and 0 when the mutual information between the two partitions equals the value expected due to chance alone.

In order to measure the gain of information about the system condition provided by the encoded representation, we evaluated the difference in $\text{AMI}$ between the latent space $z$ and the input signals $x$. Hence, $\text{AMIG}$ evaluates the encoder's capacity to find a useful representation that is informative of the system conditions and is defined as follows:
\begin{align}
    \text{AMIG}(x,z,V) = \text{AMI}(z,V) - \text{AMI}(x,V)
\end{align}

\textit{Linear Separability Gain (LSG)}: This metric evaluates the theoretical capacity of the \textit{encoder} transformation to obtain a latent space that is linearly separable in the true system states (i.e., $V$). The linear separability of the latent space (i.e., $z$ and $z_s$) is measured as the accuracy of a linear logistic classier (\textit{g}) to predict each of the system states $V$. Therefore, it uses a supervised classification problem where the true labels ($V$) are known. The gain in linear separability is then defined as accuracy gain achieved by the \textit{encoder} transformation. Hence, it evaluates if the latent representation is useful for fault segmentation. A standard logistic classifier with a linear kernel was considered. A one-vs-one strategy was assumed to reduce the original multi-class classification problem to multiple binary classification problems. The separability gain is defined as: 
\begin{align}
    \text{LSG}(z,x;V) = \text{Acc}(g(z),V) - \text{Acc}(g(x),V)
\end{align}
\textit{Mean Mutual Information (MMI)}: The mutual information for continuous target variables (i.e., $I$) is a measure of dependency between two random variables. Hence, $I(x_i,z_j)$ measures how informative each component of input signal (i.e., $x_i \;|\;  i=1, \dots, n$) is for each of the components of the latent representation (i.e., $z_j \;|\;  j=1, \dots, d$). In order to have a simple aggregated measure of an overall dependency between $x$ and $z$, we compute the mean value of mutual information map $\hat{I}(x,z)$ as follows:
\begin{align}
    \hat{I}(x,z) = \sum_{i=1}^n \sum_{j=1}^d I(x_i,z_j)
\end{align}
For autoencoder networks, we also computed the mean value of the mutual information map between the sample latent space ($z$) and each reconstructed signal  $(\bar{x})$, i.e., $I(z_i,\bar{x}_j)$. 

These three metrics for representation learning (AMIG, LSG, and $\hat{I}(x,z)$) are computed using $z=\mu$ because the encoder networks in the VAE models provide embedding representations in terms of $\mu$ and $\log\sigma^2$ and the one class network takes $\mu$ exclusively as input.

\section{Results} \label{sec:Results}
In this section, the proposed model's performance is analyzed based on three evaluation tasks: fault detection, fault segmentation, and representation learning.

\subsection{Fault detection} \label{sec:Results-fd}

Table \ref{tb:detection} reports the performance of the ten evaluated models on fault detection with respect to the three considered evaluation metrics. The different models show very different accuracy values. The proposed model, \textit{KIL-AdaVAE}, is the only model that achieves perfect accuracy. \textit{KIL-AdaVAE} provides a nearly 1.4\%  absolute improvement over the next best performing model \textit{SLE-AdaVAE} and 34\% absolute improvement over worst performing model \emph{SLE-HELM}. The supervised model \emph{SL-FF} outperforms the embedding model \emph{SLE-AE} with nearly 17\% absolute improvement. Within the embedding models, generative models based on VAE (i.e., \textit{SLE-VAE}, \textit{SLE-$\beta$-VAE} and \textit{SLE-AdaVAE}) outperform the \emph{SL-FF} discriminative model with more than a 7\% margin. Hence, these results suggest an advantage of VAE models for fault detection. Comparing $\text{SLE}$ VAE models (i.e \textit{SLE-VAE} and \textit{SLE-AdaVAE}) and $\text{SSL}$ VAE models (i.e \textit{SSL-M1-VAE} and \textit{SSL-M1-AdaVAE}), one can observe that the use of unlabeled data for training of the autoencoder network does not result in an accuracy improvement. Only when the unlabeled data is combined with the adaptive sampling (i.e., \text{SSL-M1-AdaVAE}) or with the knowledge induced learning \text{KIL-VAE}, accuracy results close to the \text{SLE} models can be achieved. All the models can clearly label the healthy class within the datasets $\{D_U, D_T\}$. Hence, the false-negative rate is zero.

%%%%%%%%%%%%%%% begin table - Comp  %%%%%%%%%%%%%%%%%%%%%%%%%%
\begin{table}[ht]
  \caption[Table caption text]{Overview of Fault Detection Results.\\ Mean values of 10 runs}
  \begin{center}
    \label{tb:detection}
    \begin{tabular}{lccc}
      \hline
      Method            & $\text{Acc}$             & $\text{TN}$                & $\text{FN}$             \\ \hline
      SL-FF             &  85.5 $\pm$ 1.6          &  84.0 $\pm$ 1.8            & 0.0 $\pm$ 0.0           \\
      SLE-AE            &  68.9 $\pm$ 0.2          &  65.6 $\pm$ 0.2            & 0.0 $\pm$ 0.0           \\
      SLE-HELM          &  65.7 $\pm$ 21.3         &  -                         & -                       \\
      SLE-VAE           &  93.3 $\pm$ 0.8          &  92.7 $\pm$ 1.6            & 0.0 $\pm$ 0.0           \\
      SLE-$\beta$-VAE   &  93.4 $\pm$ 0.1          &  92.6 $\pm$ 0.1            & 0.0 $\pm$ 0.0           \\
      SLE-AdaVAE        &  98.6 $\pm$ 0.1          &  98.5 $\pm$ 0.1            & 0.0 $\pm$ 0.0           \\
      SSL-M1-VAE        &  37.8 $\pm$ 2.5          &  31.1 $\pm$ 2.8            & 0.0 $\pm$ 0.0           \\
      SSL-M1-AdaVAE     &  93.7 $\pm$ 0.4          &  93.1 $\pm$ 0.4            & 0.0 $\pm$ 0.0           \\
      KIL-VAE           &  90.1 $\pm$ 1.3          &  89.1 $\pm$ 1.5            & 0.0 $\pm$ 0.0           \\
      KIL-AdaVAE        & \textbf{100.0 $\pm$ 0.0} & \textbf{100.0 $\pm$ 0.0}   & \textbf{0.0 $\pm$ 0.0}  \\ \hline
    \end{tabular}
  \end{center}
  \vspace{-0mm}
\end{table}
%%%%%%%%%%%%%%%% end table %%%%%%%%%%%%%%%%%%%%%%%%%%%%%%%%%%
\vspace{0mm}

\subsection{Fault segmentation} \label{sec:Results-fc}
Table \ref{tb:cluster} shows the clustering performance of the \text{OPTICS} algorithm applied to the input space ($x$) and to the latent space ($z$) of the ten evaluated models on the combined dataset $\{D_U, D_T\}$. On one hand, direct clustering in the input space identifies 14 clusters. On the other hand, the existing 18 system states (i.e., 17 fault types and the healthy one) are discovered in the latent representation of the proposed method \textit{KIL-AdaVAE}. Within the supervised with embedding models, the latent space segmentation of all the VAE models results in 18 clusters. On the contrary, for the discriminative model, \emph{SLE-AE}, only 14 clusters are discovered. Hence, these results also show an advantage of the VAE models for fault segmentation. \text{SSL} strategies lead to a reduction in the number of clusters, showing that the use of unlabeled data at training time results in a latent space that is not clearly representative of the true system states. Only when the unlabeled data is combined with a knowledge induced learning strategy (i.e., \text{KIL} models (i.e., \text{KIL-VAE} and \text{KIL-AdaVAE}), can the existing 18 clusters be distinguished. Hence, this results suggest that only through knowledge induction an good fault segmentation can be obtained.

Since the adjusted mutual information ($\text{AMI}$) evaluates how closely the segmentation proposed by the clustering algorithm replicates the true classes, the $\text{AMI}$ shows a correlation with the number of clusters. Hence, with the proposed model \textit{KIL-AdaVAE}, the cluster assignments obtained by the OPTICS algorithm in the latent space $\textbf{U}=\{U_j|j=1, \dots, R\}$ offer the closest segmentation to the true class $\textbf{V}=\{V_j|j=1, \dots, C\}$. The obtained clusters result in a nearly perfect \textit{homogeneity (h)}, indicating that its clusters contain only data points which are members of a single class. Also the \textit{completeness (c)} indicator shows a nearly perfect performance since the data points that are members of a given class $U_j$ are elements of the same cluster.  

%%%%%%%%%%%%%%% begin table - Cluster  %%%%%%%%%%%%%%%%%%%%%%%%%%
\begin{table}[ht]
  \caption[Table caption text]{Overview Unsupervised Clustering Results with OPTICS.}
  \begin{center}
    \label{tb:cluster}
    \begin{tabular}{lccccc}
      \hline
      Method            & $R$           & $\text{AMI}$  & $h$           & $c$            \\ \hline
      Input ($x$)       &  14           & 0.77          & 0.71          &  0.84          \\
      SL-FF             &  17 $\pm$ 1   & 0.90          & 0.89          &  0.91          \\
      SLE-AE            &  14           & 0.75          & 0.69          &  0.82          \\
      SLE-HELM          &   3           & -             & -             &  -             \\
      SLE-VAE           &  18           & 0.91          & 0.91          &  0.90          \\
      SLE-$\beta$-VAE   &  18           & 0.88          & 0.89          &  0.88          \\
      SLE-AdaVAE        &  18           & 0.90          & 0.91          &  0.90          \\
      SSL-M1-VAE        &  16 $\pm$ 1   & 0.78          & 0.73          &  0.83          \\ 
      SSL-M1-AdaVAE     &  17           & 0.88          & 0.87          &  0.90          \\
      KIL-VAE           &  \textbf{18}  & \textbf{0.92} & \textbf{0.93}
      & \textbf{0.92}         \\
      KIL-AdaVAE        &  \textbf{18}  & \textbf{0.92} & \textbf{0.93} & \textbf{0.92}  \\ \hline
    \end{tabular}
  \end{center}
  \vspace{-0mm}
\end{table}
%%%%%%%%%%%%%%%% end table %%%%%%%%%%%%%%%%%%%%%%%%%%%%%%%%%%

\begin{figure}[ht]
\centering
\includegraphics[width=8.7cm]{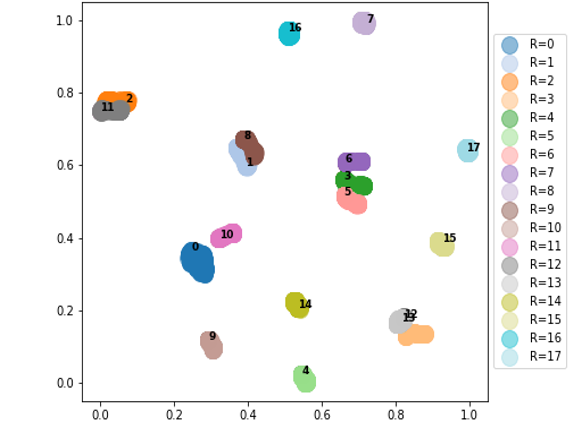}
\caption{2-D t-SNE embedding of the 8-D latent space ($z$) with perplexity 200 for \text{KIL-AdaVAE} model. Color code according to the cluster labels assigned by the OPTICS algorithm (i.e $j=0, \dots , R$ with $R=18$). The healthy class is represented with label $R=0$ (dark blue). The clustering algorithm identifies 18 clusters, therefore, each fault type and the healthy system condition is represented with a different color. Although some of the clusters overlap in the 2-D representation they are separable in the 8-D latent space.}
\label{fig:t-SNE-z}
\end{figure}

\textbf{Fault segmentation visualization}. To provide a visualization of the cluster assignments obtained with the OPTICS clustering algorithm, we perform a 2-dimensional t-Distributed Stochastic Neighbor Embedding (t-SNE) of the latent space $z$ in the combined dataset $\{D_U, D_T\}$. Figure \ref{fig:t-SNE-z} shows the resulting normalized 2-D t-SNE representation. The color code corresponds to the cluster number assigned by the OPTICS algorithm (i.e., $U_i$). t-SNE finds a lower-dimensional representation that maintains local properties of the high dimensional space. Hence, features that are close in the latent space are also close in the t-SNE representation. Therefore, since 18 clusters are identified in the eight-dimensional latent space, nearly all of them are also clearly clustered in the 2-D t-SNE space. The perplexity hyperparameter influences the cluster representation of the t-SNE. We used a perplexity value of 200 (i.e., similar to the size of the fault types) to show a very compact representation of the different system states. 
 
\subsection{Representation learning}
\label{sec:Results-fl}
The results presented in the previous sections demonstrate that the proposed knowledge induced learning method leads to an excellent performance for fault detection and segmentation. Aiming at providing a deeper understanding of how the different latent representations affect the performance of down-stream diagnostics tasks, in this section, the latent representation of each neural network model is analyzed. 

\begin{figure}[ht] 
\centering
\includegraphics[width=8cm]{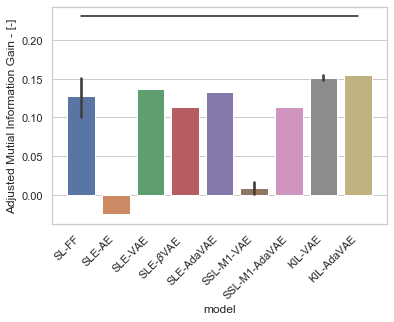}
\caption{Adjusted mutual information gain \text{AMIG} each model. The adjusted mutual information of the input space is 0.77. The horizontal black line shows the value of \text{AMIG} to achieve identical partitions to the true system states.}
\label{fig:amig}
\end{figure}

First, we evaluate how informative the encoded representation is of the true system states $\textbf{V}=\{V_j|j=1, \dots, C\}$ relative to the input space. Figure \ref{fig:amig} reports the gain of adjusted mutual information that is obtained with the cluster representation $\textbf{U}=\{U_i|i=1, \dots, R\}$ on the latent space ($z$) with respect to an alternative cluster representation $\textbf{U'}=\{U'_i|i=1, \dots, R'\}$ on the input space ($x$). The adjusted mutual information of the input space is 0.77 and, therefore, the horizontal black line shows the value of \text{AMIG} to achieve identical partitions (i.e., \text{AMIG} = 23\%). The proposed method \text{KIL-AdaVAE} provides the maximum increase in mutual information gain within the ten evaluated models. Hence, it provides the most informative latent space for fault segmentation.

\begin{figure}[ht] 
\centering
\includegraphics[width=8cm]{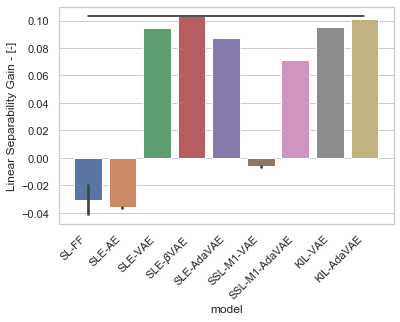}
\caption{Linear separability gain \text{LSG} of the latent space $z$ of each model relative to the input space $x$. A linear logistic classifier reaches an accuracy of 89\% predicting each of the system states $V$}
\label{fig:lsg}
\end{figure}

Extending the representation analysis, we evaluate the theoretical capacity of the \textit{encoder} transformation to obtain a representation of the input space that is linearly separable in the true system states (i.e $V$). Figure \ref{fig:lsg} shows the gain of linear separability that the latent space provides relative to the input space for each model. A logistic classifier with a one-class versus the rest (one-vs-rest) approach results in a 89\% accuracy and, therefore, the horizontal black line shows the value of \text{LSG} to achieve perfect accuracy (i.e. \text{LSG} = 11\%). The embedding of the proposed approach results in a 10\% gain. Hence, a linear decision boundary can nearly perfectly separate each class versus the rest. 

\textbf{Information Preservation}. The previous analysis shows that the embedding representation of the input signals is very informative of the true class and is linearly separable. We focus now on the analysis of the information preservation through the bottleneck.

Table \ref{tb:MMI} reports the mean value of the mutual information map between the input space and its embedding representation $\hat{I}(x,z)$. For the autoencoder configurations, it also provides the mean value of the mutual information map between the \textit{sample} latent space ($z$) and each of the reconstructed signals $\hat{I}(z,\bar{x})$. The autoencoder models show similar values of $\hat{I}(x,z)$, which are higher than for the \text{SL-FF} model. The latent space of the supervised model is taken at the second hidden layer of the network, and at this depth, the information about the input is quickly reduced. Traditional VAE models show nearly zero mutual information between the \textit{sample} latent space ($z$) and each of the reconstructed signals ($\bar{x}$) since the sampled representation ($z$) is clearly dominated by the random Gaussian noise ($\epsilon$). In contrast, the proposed sampling method (i.e., \text{SLE-AdaVAE}, \text{SSL-M1-AdaVAE} and \text{KIL-AdaVAE}) increases the amount of information between the sampled representation and the reconstructed signals. This also corresponds with a less noisy reconstruction signal. The proposed sampling method balances reconstruction and regularization of the \text{ELBO}. Therefore, the values $\hat{I}(z,\bar{x})$ on the sampled representation are 40\% smaller than traditional autoencoders (\text{SLE-AE} and \text{SLE-HELM}).

%%%%%%%%%%%%%%% begin table - Representation  %%%%%%%%%%%%%%%%%%%%%%%%%%
\begin{table}[ht]
  \caption[Table caption text]{Mean Mutual Information between input and latent space $\hat{I}(x,z)$ and between latent space and reconstructed input $\hat{I}(z,\hat{x})$}
  \begin{center}
    \label{tb:MMI}
    \begin{tabular}{lcc}
      \hline
      Method            & $\hat{I}(x,z)$   &  $\hat{I}(z,\bar{x})$    \\   \hline
      SL-FF             &  0.49            &  -                       \\
      SLE-AE            &  2.90            &  3.56                    \\
      SLE-HELM          &  3.52            &  4.51                    \\
      SLE-VAE           &  3.04            &  0.09                    \\
      SLE-$\beta$-VAE   &  3.00            &  0.06                    \\
      SLE-AdaVAE        &  3.13            &  1.31                    \\
      SSL-M1-VAE        &  3.10            &  0.10                    \\
      SSL-M1-AdaVAE     &  3.06            &  1.51                    \\    
      KIL-VAE           &  2.55            &  0.03                    \\
      KIL-AdaVAE        &  2.88            &  0.69                    \\
      \hline
    \end{tabular}
  \end{center}
  \vspace{-5mm}
\end{table}
%%%%%%%%%%%%%%%% end table %%%%%%%%%%%%%%%%%%%%%%%%%%%%%%%%%%
\subsection{Ablation study  - Hyperparamenter variation} \label{sec:betaVAE}
\textbf{Effect of the $\gamma$-parameter}.  Table \ref{tb:KIL-AdaVAE} reports the impact of increasing the values of $\gamma$ for the \text{KIL-AdaVAE} model in detection accuracy, number of clusters ($R$) and the adjusted mutual information (AMI) of the latent space. We can observe that the \text{KIL-AdaVAE} method is fairly robust to a large range of values of $\gamma$. 

%%%%%%%%%%%%%%% begin table - KIL-AdaVAE  %%%%%%%%%%%%%%%%%%%%%%%%%%%
\begin{table}[ht]
  \caption[Table caption text]{Effect of $\gamma$ in \text{KIL-AdaVAE} model}
  \begin{center}
    \label{tb:KIL-AdaVAE}
    \begin{tabular}{lccc}
      \hline
      $\gamma$ & $\text{Acc}$   & $R$   & AMI   \\  \hline
      $10$     &  99.7          & 17    & 0.86  \\
      $12$     &  100.0         & 18    & 0.92  \\
      $15$     &  99.7          & 18    & 0.93  \\
      $17$     &  99.0          & 18    & 0.93  \\
      \hline
    \end{tabular}
  \end{center}
%  \vspace{-5mm}
\end{table}
%%%%%%%%%%%%%%%% end table %%%%%%%%%%%%%%%%%%%%%%%%%%%%%%%%%%%%%^%

\textbf{Effect of the $\beta$-parameter - Avoiding the posterior collapse.} Table \ref{tb:betaVAE} and Table \ref{tb:adaVAE} report the impact of increasing the values of $\beta$ for the models \text{SLE-$\beta$-VAE} and \text{SLE-AdaVAE} in detection accuracy ($\text{Acc}$), mean mutual information $\hat{I}(x,\mu)$, mean mutual information $\hat{I}(z,\bar{x})$, reconstruction loss $||x-\bar{x}||_{2}$ and optimisation loss $\mathcal{L}_{\beta \text{-VAE}}$. As $\beta$ increases, detection accuracy of \text{SLE-$\beta$-VAE} model rapidly degrades. We also observe a decrease in the information preserved by the latent representation. This information loss is manifested in the increase of the reconstruction loss ($\ell_{2}$-loss). In contrast, the \text{SLE-AdaVAE} is able to maintain very low reconstruction error while keeping the $D_{\text{KL}}$ term, and hence the total loss, small. Therefore, the \text{SLE-AdaVAE} model avoids the posterior collapse at high $\beta$ and as a result is able to maintain a high detection accuracy.

%%%%%%%%%%%%%%% begin table - SLE-BetaVAE  %%%%%%%%%%%%%%%%%%%%%%%%%%
\begin{table}[ht]
  \caption[Table caption text]{Effect of $\beta$ in $\beta$-VAE model}
  \begin{center}
    \label{tb:betaVAE}
    \begin{tabular}{lccccc}
      \hline
      $\beta$ & $\text{Acc}$      & $\hat{I}(x,\mu)$  & $\hat{I}(z,\bar{x})$ & $\ell_{2}$-loss  & $\mathcal{L}_{\beta \text{-VAE}}$ \\  \hline
      $1$     &  93.3 $\pm$ 1.0   &  3.05           & 0.09                 & 0.028            & 1.42                              \\
      $5$     &  93.4 $\pm$ 0.7   &  3.00           & 0.06                 & 0.240            & 3.73                              \\
      $9$     &  69.3 $\pm$ 3.2   &  2.83           & 0.04                 & 0.292            & 3.80                              \\
      $13$    &  68.6 $\pm$ 6.7   &  2.77           & 0.03                 & 0.292            & 3.80                              \\
      \hline
    \end{tabular}
  \end{center}
%  \vspace{-5mm}
\end{table}
%%%%%%%%%%%%%%%% end table %%%%%%%%%%%%%%%%%%%%%%%%%%%%%%%%%%%%%%%
%
%%%%%%%%%%%%%%% begin table - SLE-AdaVAE  %%%%%%%%%%%%%%%%%%%%%%%%%%%
\begin{table}[ht]
  \caption[Table caption text]{Effect of $\beta$ in \text{AdaVAE} model}
  \begin{center}
    \label{tb:adaVAE}
    \begin{tabular}{lccccc}
      \hline
      $\beta$ & $\text{Acc}$      & $\hat{I}(x,\mu)$  & $\hat{I}(z,\bar{x})$ & $\ell_{2}$-loss  & $\mathcal{L}_{\beta \text{-AdaVAE}}$ \\  \hline
      $1$     &  93.7 $\pm$ 1.8   &  3.05           &  1.34                & 0.0006           & 0.022                          \\
      $5$     &  98.6 $\pm$ 0.1   &  3.13           &  1.32                & 0.0008           & 0.032                          \\
      $9$     &  92.5 $\pm$ 0.9   &  3.15           &  1.28                & 0.0011           & 0.040                          \\
      $13$    &  96.0 $\pm$ 1.3   &  3.04           &  1.19                & 0.0013           & 0.046                          \\
      \hline
    \end{tabular}
  \end{center}
%  \vspace{-5mm}
\end{table}
%%%%%%%%%%%%%%%% end table %%%%%%%%%%%%%%%%%%%%%%%%%%%%%%%%%%%%%^%

\section{Conclusions and Future Work} \label{sec:Conclusion}
In this work, we proposed $\text{KIL-AdaVAE}$, a knowledge induced learning method with adaptive sampling variational autoencoder for accurate open-set diagnosis of complex systems. The performance of the proposed method was evaluated using a new dataset generated with the Advanced Geared Turbofan 30,000 (AGTF30) dynamical model. \text{KIL-AdaVAE} significantly outperformed other methods in supervised and semi-supervised settings providing a 100\% fault detection accuracy and a very good fault segmentation. Through an extensive feature representation analysis, we demonstrated that its excellent performance is rooted in the fact that the resulting latent representation of the sensor readings i.e., $z$ is not only very informative about the true system conditions but is also linearly separable with respect to those conditions (healthy system conditions and all observed fault types). More importantly, we demonstrate that \text{KIL-AdaVAE} reaches this discriminative latent representation by only having access to the healthy system conditions.

Although the current problem formulation leads to an excellent result, from a theoretical perspective, the proposed adaptive sampling implies a deviation from the Gaussian prior generally assumed in the distribution of the variational approximate posterior that has not been considered in the $\text{KL}$ term to keep an analytical close expression. A reformulation of the \text{KL} term to fit with the VAE framework remains, therefore, as a future line of research. Finally, from the prognostics and health management perspective, the potential of the proposed solution in a setting with more extensive operating conditions is a subject for further research.

\section*{Acknowledgment}
This research was funded by the Swiss National Science Foundation (SNSF) Grant no. PP00P2\_176878.

\bibliographystyle{unsrt} 
\bibliography{reference}

\begin{thebibliography}{10}

\bibitem{Yin2016}
Zuyu Yin and Jian Hou.
\newblock {Recent advances on SVM based fault diagnosis and process monitoring
  in complicated industrial processes}.
\newblock {\em Neurocomputing}, 174:643--650, jan 2016.

\bibitem{Jing2017}
Luyang Jing, Ming Zhao, Pin Li, and Xiaoqiang Xu.
\newblock {A convolutional neural network based feature learning and fault
  diagnosis method for the condition monitoring of gearbox}.
\newblock {\em Measurement}, 111:1--10, dec 2017.

\bibitem{Wang2018StackedSA}
Ying Wang, Meiqin Liu, Zhejing Bao, and Senlin Zhang.
\newblock Stacked sparse autoencoder with pca and svm for data-based line trip
  fault diagnosis in power systems.
\newblock {\em Neural Computing and Applications}, 31:6719--6731, 2018.

\bibitem{Lei2019}
Jinhao Lei, Chao Liu, and Dongxiang Jiang.
\newblock {Fault diagnosis of wind turbine based on Long Short-term memory
  networks}.
\newblock {\em Renewable Energy}, 133:422--432, apr 2019.

\bibitem{scheirer2012toward}
Walter~J Scheirer, Anderson de~Rezende~Rocha, Archana Sapkota, and Terrance~E
  Boult.
\newblock Toward open set recognition.
\newblock {\em IEEE transactions on pattern analysis and machine intelligence},
  35(7):1757--1772, 2012.

\bibitem{Moya1996}
Mary~M. Moya and Don~R. Hush.
\newblock {Network constraints and multi-objective optimization for one-class
  classification}.
\newblock {\em Neural Networks}, 9(3):463--474, apr 1996.

\bibitem{Scholkopf2000}
Bernhard Sch{\"{o}}lkopf, Robert Williamson, Alex Smola, John Shawe-Taylor, and
  John Piatt.
\newblock {Support vector method for novelty detection}.
\newblock In {\em Advances in Neural Information Processing Systems}, pages
  582--588, 2000.

\bibitem{pmlr-v80-ruff18a}
Lukas Ruff, Robert Vandermeulen, Nico Goernitz, Lucas Deecke, Shoaib~Ahmed
  Siddiqui, Alexander Binder, Emmanuel M{\"u}ller, and Marius Kloft.
\newblock Deep one-class classification.
\newblock In Jennifer Dy and Andreas Krause, editors, {\em Proceedings of the
  35th International Conference on Machine Learning}, volume~80 of {\em
  Proceedings of Machine Learning Research}, pages 4393--4402,
  Stockholmsmässan, Stockholm Sweden, 10--15 Jul 2018. PMLR.

\bibitem{chalapathy2018anomaly}
Raghavendra Chalapathy, Aditya~Krishna Menon, and Sanjay Chawla.
\newblock Anomaly detection using one-class neural networks.
\newblock {\em CoRR}, abs/1802.06360, 2018.

\bibitem{Fernandez-Francos2013}
Diego Fern{\'{a}}ndez-Francos, David Mart{\'{e}}nez-Rego, Oscar
  Fontenla-Romero, and Amparo Alonso-Betanzos.
\newblock {Automatic bearing fault diagnosis based on one-class m-SVM}.
\newblock {\em Computers and Industrial Engineering}, 64(1):357--365, jan 2013.

\bibitem{Michau2017}
Gabriel Michau, Thomas Palm{\'{e}}, and Olga Fink.
\newblock {Deep Feature Learning Network for Fault Detection and Isolation}.
\newblock {\em Conference of the PHM Society}, 8(012):1--11, 2017.

\bibitem{AriasChao2019}
Manuel Arias~Chao, Chetan Kulkarni, Kai Goebel, and Olga Fink.
\newblock Hybrid deep fault detection and isolation: Combining deep neural
  networks and system performance models.
\newblock {\em International Journal of Prognostics and Health Management},
  10:033, 2019.

\bibitem{Song2017AHS}
Hongchao Song, Zhuqing Jiang, Aidong Men, and Bo~Yang.
\newblock A hybrid semi-supervised anomaly detection model for high-dimensional
  data.
\newblock {\em Computational Intelligence and Neuroscience}, 2017, 2017.

\bibitem{Akay2018GANomalySA}
Samet Akcay, Amir~Atapour Abarghouei, and Toby~P. Breckon.
\newblock Ganomaly: Semi-supervised anomaly detection via adversarial training.
\newblock {\em CoRR}, abs/1805.06725, 2018.

\bibitem{Chalapathy2019}
Raghavendra Chalapathy and Sanjay Chawla.
\newblock Deep learning for anomaly detection: {A} survey.
\newblock {\em CoRR}, abs/1901.03407, 2019.

\bibitem{Detroja2006}
K.P. Detroja, R.D. Gudi, and S.C. Patwardhan.
\newblock {A possibilistic clustering approach to novel fault detection and
  isolation}.
\newblock {\em Journal of Process Control}, 16(10):1055--1073, dec 2006.

\bibitem{HuDi2012}
Di~Hu, Ali Sarosh, and Yun-Feng Dong.
\newblock {A novel KFCM based fault diagnosis method for unknown faults in
  satellite reaction wheels}.
\newblock {\em ISA Transactions}, 51(2):309--316, mar 2012.

\bibitem{Li2014}
Chaoshun Li and Jianzhong Zhou.
\newblock {Semi-supervised weighted kernel clustering based on gravitational
  search for fault diagnosis}.
\newblock {\em ISA Transactions}, 53(5):1534--1543, sep 2014.

\bibitem{Du2014a}
Zhimin Du, Bo~Fan, Xinqiao Jin, and Jinlei Chi.
\newblock {Fault detection and diagnosis for buildings and HVAC systems using
  combined neural networks and subtractive clustering analysis}.
\newblock {\em Building and Environment}, 73:1--11, mar 2014.

\bibitem{Costa2015}
Bruno Sielly~Jales Costa, Plamen~Parvanov Angelov, and Luiz~Affonso Guedes.
\newblock {Fully unsupervised fault detection and identification based on
  recursive density estimation and self-evolving cloud-based classifier}.
\newblock {\em Neurocomputing}, 150:289--303, feb 2015.

\bibitem{Zhu2017}
Kedong Zhu, Fei Mei, and Jianyong Zheng.
\newblock {Adaptive fault diagnosis of HVCBs based on P-SVDD and P-KFCM}.
\newblock {\em Neurocomputing}, 240:127--136, may 2017.

\bibitem{Ratner2019}
Alex Ratner, Varma Paroma, Braden Hancock, and Chris R{\'{e}}.
\newblock {Weak Supervision: A New Programming Paradigm for Machine Learning |
  SAIL Blog}, 2019.

\bibitem{Jiang2013}
Li~Jiang, Jianping Xuan, and Tielin Shi.
\newblock {Feature extraction based on semi-supervised kernel Marginal Fisher
  analysis and its application in bearing fault diagnosis}.
\newblock {\em Mechanical Systems and Signal Processing}, 41(1-2):113--126, dec
  2013.

\bibitem{Hu2017}
Yang Hu, Piero Baraldi, Francesco {Di Maio}, and Enrico Zio.
\newblock {A Systematic Semi-Supervised Self-adaptable Fault Diagnostics
  approach in an evolving environment}.
\newblock {\em Mechanical Systems and Signal Processing}, 88:413--427, may
  2017.

\bibitem{Chen2018}
Xinan Chen, Zhipeng Wang, Zhe Zhang, Limin Jia, and Yong Qin.
\newblock {A Semi-Supervised Approach to Bearing Fault Diagnosis under Variable
  Conditions towards Imbalanced Unlabeled Data}.
\newblock {\em Sensors}, 18(7):2097, jun 2018.

\bibitem{Razavi-Far2019}
Roozbeh Razavi-Far, Ehsan Hallaji, Maryam Farajzadeh-Zanjani, and Mehrdad Saif.
\newblock {A semi-supervised diagnostic framework based on the surface
  estimation of faulty distributions}.
\newblock {\em IEEE Transactions on Industrial Informatics}, 15(3):1277--1286,
  mar 2019.

\bibitem{Chapelle}
Olivier Chapelle, Bernhard Schlkopf, and Alexander Zien.
\newblock {\em Semi-Supervised Learning}.
\newblock The MIT Press, 1st edition, 2010.

\bibitem{Ruff2020Deep}
Lukas Ruff, Robert~A. Vandermeulen, Nico Görnitz, Alexander Binder, Emmanuel
  Müller, Klaus-Robert Müller, and Marius Kloft.
\newblock Deep semi-supervised anomaly detection.
\newblock In {\em International Conference on Learning Representations}, 2020.

\bibitem{Bengio2013}
Yoshua Bengio, Aaron Courville, and Pascal Vincent.
\newblock {Representation learning: A review and new perspectives}.
\newblock {\em IEEE Transactions on Pattern Analysis and Machine Intelligence},
  35(8):1798--1828, jun 2013.

\bibitem{Kingma2013}
Diederik~P. Kingma and Max Welling.
\newblock Auto-encoding variational bayes.
\newblock In Yoshua Bengio and Yann LeCun, editors, {\em 2nd International
  Conference on Learning Representations, {ICLR} 2014, Banff, AB, Canada, April
  14-16, 2014, Conference Track Proceedings}, 2014.

\bibitem{Higgins2017}
Irina Higgins, Lo{\"{\i}}c Matthey, Arka Pal, Christopher Burgess, Xavier
  Glorot, Matthew Botvinick, Shakir Mohamed, and Alexander Lerchner.
\newblock beta-vae: Learning basic visual concepts with a constrained
  variational framework.
\newblock In {\em 5th International Conference on Learning Representations,
  {ICLR} 2017, Toulon, France, April 24-26, 2017, Conference Track
  Proceedings}. OpenReview.net, 2017.

\bibitem{pmlr-v80-kim18b}
Hyunjik Kim and Andriy Mnih.
\newblock Disentangling by factorising.
\newblock In Jennifer Dy and Andreas Krause, editors, {\em Proceedings of the
  35th International Conference on Machine Learning}, volume~80 of {\em
  Proceedings of Machine Learning Research}, pages 2649--2658,
  Stockholmsmässan, Stockholm Sweden, 10--15 Jul 2018. PMLR.

\bibitem{Locatello2019}
Francesco Locatello, Stefan Bauer, Mario Lucic, Gunnar R{\"{a}}tsch, Sylvain
  Gelly, Bernhard Sch{\"{o}}lkopf, and Olivier Bachem.
\newblock Challenging common assumptions in the unsupervised learning of
  disentangled representations.
\newblock In {\em Reproducibility in Machine Learning, {ICLR} 2019 Workshop,
  New Orleans, Louisiana, United States, May 6, 2019}. OpenReview.net, 2019.

\bibitem{Chapman2017}
Jeffryes~W. Chapman and Jonathan~S. Litt.
\newblock {Control Design for an Advanced Geared Turbofan Engine}.
\newblock In {\em 53rd AIAA/SAE/ASEE Joint Propulsion Conference}, Reston,
  Virginia, jul 2017. American Institute of Aeronautics and Astronautics.

\bibitem{Borghesi2019}
Andrea Borghesi, Andrea Bartolini, Michele Lombardi, Michela Milano, and Luca
  Benini.
\newblock {A semisupervised autoencoder-based approach for anomaly detection in
  high performance computing systems}.
\newblock {\em Engineering Applications of Artificial Intelligence},
  85:634--644, oct 2019.

\bibitem{Gornitz2013}
Nico G\"{o}rnitz, Marius Kloft, Konrad Rieck, and Ulf Brefeld.
\newblock Toward supervised anomaly detection.
\newblock {\em J. Artif. Int. Res.}, 46(1):235–262, January 2013.

\bibitem{Kiran2018}
B.~Kiran, Dilip Thomas, and Ranjith Parakkal.
\newblock {An Overview of Deep Learning Based Methods for Unsupervised and
  Semi-Supervised Anomaly Detection in Videos}.
\newblock {\em Journal of Imaging}, 4(2):36, feb 2018.

\bibitem{Min2018}
Erxue Min, Jun Long, Qiang Liu, Jianjing Cui, Zhiping Cai, and Junbo Ma.
\newblock Su-ids: A semi-supervised and unsupervised framework for network
  intrusion detection.
\newblock In Xingming Sun, Zhaoqing Pan, and Elisa Bertino, editors, {\em Cloud
  Computing and Security}, pages 322--334, Cham, 2018. Springer International
  Publishing.

\bibitem{KingmaSemi}
Diederik~P. Kingma, Danilo~J. Rezende, Shakir Mohamed, and Max Welling.
\newblock {Semi-supervised learning with deep generative models}.
\newblock In {\em Advances in Neural Information Processing Systems}, volume~4,
  pages 3581--3589, jun 2014.

\bibitem{Bekker2020}
Jessa Bekker and Jesse Davis.
\newblock {Learning from positive and unlabeled data: a survey}.
\newblock {\em Machine Learning}, 109(4):719--760, apr 2020.

\bibitem{An2015VariationalAB}
Jinwon An and Sungzoon Cho.
\newblock Variational autoencoder based anomaly detection using reconstruction
  probability.
\newblock In {\em SNU Data Mining Center, 2015-2 Special Lecture on IE}, 2015.

\bibitem{RIBEIRO201813}
Manassés Ribeiro, André~Eugênio Lazzaretti, and Heitor~Silvério Lopes.
\newblock A study of deep convolutional auto-encoders for anomaly detection in
  videos.
\newblock {\em Pattern Recognition Letters}, 105:13 -- 22, 2018.
\newblock Machine Learning and Applications in Artificial Intelligence.

\bibitem{Linsker1988}
R.~{Linsker}.
\newblock Self-organization in a perceptual network.
\newblock {\em Computer}, 21(3):105--117, 1988.

\bibitem{Bell1995}
Anthony~J. Bell and Terrence~J. Sejnowski.
\newblock An information-maximization approach to blind separation and blind
  deconvolution.
\newblock {\em Neural Comput.}, 7(6):1129–1159, November 1995.

\bibitem{hjelm2018learning}
R.~Devon Hjelm, Alex Fedorov, Samuel Lavoie{-}Marchildon, Karan Grewal, Philip
  Bachman, Adam Trischler, and Yoshua Bengio.
\newblock Learning deep representations by mutual information estimation and
  maximization.
\newblock In {\em 7th International Conference on Learning Representations,
  {ICLR} 2019, New Orleans, LA, USA, May 6-9, 2019}. OpenReview.net, 2019.

\bibitem{Zhang2020}
Yuyan Zhang, Xinyu Li, Liang Gao, Wen Chen, and Peigen Li.
\newblock {Ensemble deep contractive auto-encoders for intelligent fault
  diagnosis of machines under noisy environment}.
\newblock {\em Knowledge-Based Systems}, 196:105764, may 2020.

\bibitem{MENG2018448}
Zong Meng, Xuyang Zhan, Jing Li, and Zuozhou Pan.
\newblock An enhancement denoising autoencoder for rolling bearing fault
  diagnosis.
\newblock {\em Measurement}, 130:448 -- 454, 2018.

\bibitem{SUN2016171}
Wenjun Sun, Siyu Shao, Rui Zhao, Ruqiang Yan, Xingwu Zhang, and Xuefeng Chen.
\newblock A sparse auto-encoder-based deep neural network approach for
  induction motor faults classification.
\newblock {\em Measurement}, 89:171 -- 178, 2016.

\bibitem{SHAO2018278}
Haidong Shao, Hongkai Jiang, Ying Lin, and Xingqiu Li.
\newblock A novel method for intelligent fault diagnosis of rolling bearings
  using ensemble deep auto-encoders.
\newblock {\em Mechanical Systems and Signal Processing}, 102:278 -- 297, 2018.

\bibitem{XU20181}
Lin Xu, Maoyong Cao, Baoye Song, Jiansheng Zhang, Yurong Liu, and Fuad~E.
  Alsaadi.
\newblock Open-circuit fault diagnosis of power rectifier using sparse
  autoencoder based deep neural network.
\newblock {\em Neurocomputing}, 311:1 -- 10, 2018.

\bibitem{Locatello2020}
Francesco Locatello, Ben Poole, Gunnar R{\"{a}}tsch, Bernhard Sch{\"{o}}lkopf,
  Olivier Bachem, and Michael Tschannen.
\newblock Weakly-supervised disentanglement without compromises.
\newblock {\em CoRR}, abs/2002.02886, 2020.

\bibitem{Doersch2016}
Carl Doersch.
\newblock Tutorial on variational autoencoders.
\newblock {\em ArXiv}, abs/1606.05908, 2016.

\bibitem{Zhao2019}
Shengjia Zhao, Jiaming Song, and Stefano Ermon.
\newblock {InfoVAE: Balancing Learning and Inference in Variational
  Autoencoders}.
\newblock {\em Proceedings of the AAAI Conference on Artificial Intelligence},
  33:5885--5892, jul 2019.

\bibitem{Ahmed1989}
N.~A. {Ahmed} and D.~V. {Gokhale}.
\newblock Entropy expressions and their estimators for multivariate
  distributions.
\newblock {\em IEEE Transactions on Information Theory}, 35(3):688--692, 1989.

\bibitem{Alemi2018FixingAB}
Alexander~A. Alemi, Ben Poole, Ian Fischer, Joshua~V. Dillon, Rif~A. Saurous,
  and Kevin Murphy.
\newblock Fixing a broken {ELBO}.
\newblock In Jennifer~G. Dy and Andreas Krause, editors, {\em Proceedings of
  the 35th International Conference on Machine Learning, {ICML} 2018,
  Stockholmsm{\"{a}}ssan, Stockholm, Sweden, July 10-15, 2018}, volume~80 of
  {\em Proceedings of Machine Learning Research}, pages 159--168. {PMLR}, 2018.

\bibitem{Ester1996}
X~{Ester, M., Kriegel, H. P., Sander, J., {\&} Xu}.
\newblock {A Density-Based Algorithm for Discovering Clusters in Large Spatial
  Databases with Noise}.
\newblock {\em Kdd}, 96(34):226--231, 1996.

\bibitem{VanDerMaaten2008}
Laurens {Van Der Maaten} and Geoffrey Hinton.
\newblock {Visualizing Data using t-SNE}.
\newblock {\em Journal of Machine Learning Research}, 9:2579--2605, 2008.

\bibitem{DASHlink}
{DASHlink - Flight Data For Tail 687}, 2012.

\bibitem{Kingma2014Adam}
Diederik~P Kingma and Jimmy~Lei Ba.
\newblock {Adam: A method for stochastic optimization}.
\newblock In {\em 3rd International Conference on Learning Representations,
  ICLR 2015 - Conference Track Proceedings}, 2015.

\bibitem{Glorot}
Xavier Glorot and Yoshua Bengio.
\newblock Understanding the difficulty of training deep feedforward neural
  networks.
\newblock In Yee~Whye Teh and Mike Titterington, editors, {\em Proceedings of
  the Thirteenth International Conference on Artificial Intelligence and
  Statistics}, volume~9 of {\em Proceedings of Machine Learning Research},
  pages 249--256, Chia Laguna Resort, Sardinia, Italy, 13--15 May 2010. PMLR.

\bibitem{XuanVinh2010}
Nguyen {Xuan Vinh}, Julien Epps, and James Bailey.
\newblock {Information Theoretic Measures for Clusterings Comparison: Variants,
  Properties, Normalization and Correction for Chance}.
\newblock {\em Journal of Machine Learning Research}, 11:2837--2854, 2010.

\end{thebibliography}

\end{document}